\documentclass[10pt,twocolumn,letterpaper]{article}
\usepackage{iccv}
\usepackage{times}
\usepackage{epsfig}
\usepackage{graphicx}
\usepackage{amsmath}
\usepackage{amssymb}

\usepackage{bm}
\usepackage[flushleft]{threeparttable}

\newcommand{\tablestyle}[2]{\setlength{\tabcolsep}{#1}\renewcommand{\arraystretch}{#2}\centering\footnotesize}
\newlength\savewidth\newcommand\shline{\noalign{\global\savewidth\arrayrulewidth
		\global\arrayrulewidth 1pt}\hline\noalign{\global\arrayrulewidth\savewidth}}
\usepackage[table]{xcolor}

\usepackage{multirow}
\usepackage{mathtools}
\usepackage{array}
\usepackage{booktabs}
\usepackage{colortbl}
\usepackage{arydshln}
\usepackage{multicol}
\usepackage{makecell}
\usepackage{balance}

\usepackage{amsmath,amsfonts,bm}

\def\eqref#1{equation~\ref{#1}}

\def\1{\bm{1}}

\def\rva{{\mathbf{a}}}
\def\rvb{{\mathbf{b}}}

\def\rvk{{\mathbf{k}}}

\def\rvm{{\mathbf{m}}}

\def\rvp{{\mathbf{p}}}

\def\rvw{{\mathbf{w}}}

\def\m1{{\bm{1}}}

\DeclareMathAlphabet{\mathsfit}{\encodingdefault}{\sfdefault}{m}{sl}
\SetMathAlphabet{\mathsfit}{bold}{\encodingdefault}{\sfdefault}{bx}{n}

\DeclareMathOperator*{\argmin}{arg\,min}

\definecolor{Gray}{gray}{0.5}

\definecolor{Highlight}{HTML}{39b54a}  
\definecolor{Modify}{HTML}{2240F0}  
\newcommand{\cgaphl}[2]{
	\fontsize{6pt}{1em}\selectfont{\textcolor{Highlight}{(${#1}$\textbf{#2})}}
}
\newcommand{\hl}[1]{\textcolor{Highlight}{#1}}
\newcommand{\modify}[1]{\textcolor{Modify}{#1}}
\definecolor{Highlight}{HTML}{3DA600}  
\newcommand{\cgaphltimes}[2]{
	\fontsize{5pt}{1em}\selectfont{\textcolor{Highlight}{(\textbf{#1}{#2})}}
}

\newcommand{\cgaphlsupp}[2]{
	\fontsize{5pt}{1em}\selectfont{\textcolor{Highlight}{(\textbf{#1}{#2})}}
}

\newcommand*{\affaddr}[1]{#1} 
\newcommand*{\affmark}[1][*]{\textsuperscript{#1}}

\makeatletter
\newcommand{\printfnsymbol}[1]{\textsuperscript{\@fnsymbol{#1}}}
\makeatother

\usepackage{algorithm}
\PassOptionsToPackage{noend}{algpseudocode}
\usepackage{algpseudocode}

\errorcontextlines\maxdimen

\makeatletter
\newcommand*{\algrule}[1][\algorithmicindent]{\makebox[#1][l]{\hspace*{.5em}\vrule height .75\baselineskip depth .25\baselineskip}}%

\newcount\ALG@printindent@tempcnta
\def\ALG@printindent{%
    \ifnum \theALG@nested>0
        \ifx\ALG@text\ALG@x@notext
            \addvspace{-3pt}
        \else
            \unskip
            \ALG@printindent@tempcnta=1
            \loop
                \algrule[\csname ALG@ind@\the\ALG@printindent@tempcnta\endcsname]%
                \advance \ALG@printindent@tempcnta 1
            \ifnum \ALG@printindent@tempcnta<\numexpr\theALG@nested+1\relax
            \repeat
        \fi
    \fi
    }%
\usepackage{etoolbox}
\patchcmd{\ALG@doentity}{\noindent\hskip\ALG@tlm}{\ALG@printindent}{}{\errmessage{failed to patch}}
\makeatother

\usepackage[pagebackref=true,breaklinks=true,letterpaper=true,colorlinks,bookmarks=false]{hyperref}

\iccvfinalcopy 

\def\iccvPaperID{6722} 

\ificcvfinal\fi

\begin{document}

\title{Sub-bit Neural Networks: Learning to Compress and Accelerate\\Binary Neural Networks
}

\author{
	Yikai Wang\affmark[1]\thanks{This research was done when Yikai Wang was an intern at Intel Labs China, supervised by Anbang Yao who is responsible for correspondence.}\quad$\;$ Yi Yang\affmark[2]\quad$\;$ Fuchun Sun\affmark[1]\quad$\,$ Anbang Yao\affmark[2]\\
	\affaddr{\affmark[1]Beijing National Research Center for Information Science and Technology$\,$(BNRist),\\ State Key Lab on Intelligent Technology and Systems,\\ Department of Computer Science and Technology, Tsinghua University}\quad
	\affaddr{\affmark[2]Intel Corporation}\\
	\tt\small{\{wangyk17@mails., fcsun@\}tsinghua.edu.cn, \{yi.b.yang, anbang.yao\}@intel.com}\\
}

\maketitle
\ificcvfinal\thispagestyle{empty}\fi

\begin{abstract}
In the low-bit quantization field, training Binary Neural Networks (BNNs) is the extreme solution to ease the deployment of deep models on resource-constrained devices, having the lowest storage cost and significantly cheaper bit-wise operations compared to 32-bit floating-point counterparts. In this paper, we introduce Sub-bit Neural Networks (SNNs), a new type of binary quantization design tailored to compress and accelerate BNNs. SNNs are inspired by an empirical observation, showing that binary kernels learnt at convolutional layers of a BNN model are likely to be distributed over kernel subsets. As a result, unlike existing methods that binarize weights one by one, SNNs are trained with a kernel-aware optimization framework, which exploits binary quantization in the fine-grained convolutional kernel space. Specifically, our method includes a random sampling step generating layer-specific subsets of the kernel space, and a refinement step learning to adjust these subsets of binary kernels via optimization. Experiments on visual recognition benchmarks and the hardware deployment on FPGA validate the great potentials of SNNs. For instance, on ImageNet, SNNs of ResNet-18$/$ResNet-34 with 0.56-bit weights achieve 3.13$/$3.33$\times$ runtime speed-up and 1.8$\times$ compression over conventional BNNs with moderate drops in recognition accuracy. Promising results are also obtained when applying SNNs to binarize both weights and activations. Our code is available at \url{https://github.com/yikaiw/SNN}.
\end{abstract}

\begin{figure}[t]
\centering
\hskip-0.3em
\includegraphics[scale=0.49]{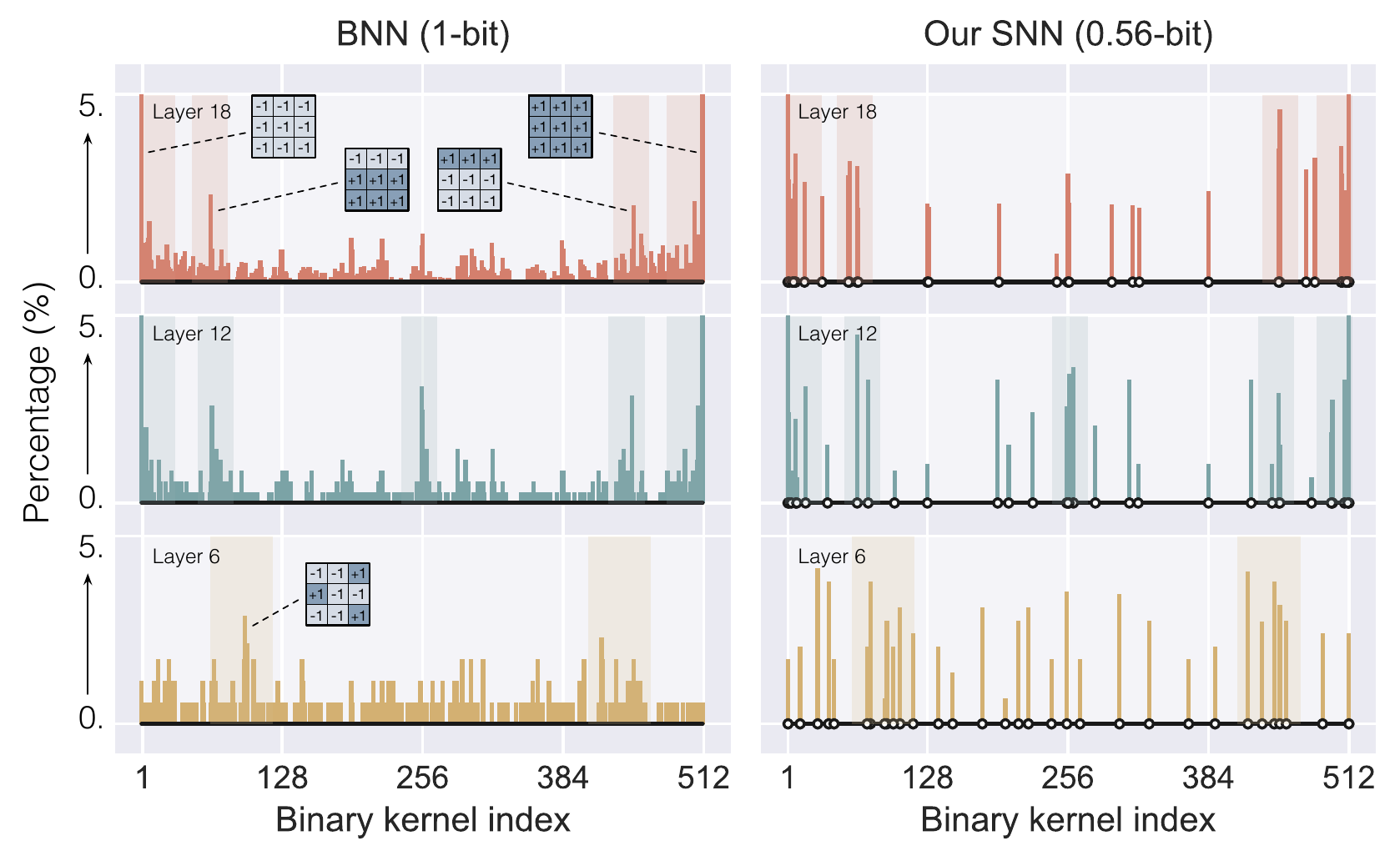}
\caption{Frequencies of different binary kernels in each layer, collected on well-trained ResNet-$20$ models. \textbf{Left:} a standard BNN model, with $1$-bit per weight, which converts each floating-point kernel to one of the $512$ binary kernels.  \textbf{Right:} the proposed SNN model, with $0.56$-bit per weight. Instead of using $512$ binary kernels, $0.56$-bit SNN only adopts $32$ binary kernels for each layer, achieving larger compression and acceleration ratios than BNN.}
\label{pic:intro}
\vskip-0.1in
\end{figure}

\section{Introduction}
To enable easy deployment of Convolutional Neural Networks (CNNs) on mobile devices, many attempts are devoted to improving network efficiency, \emph{e.g.}, reducing the  storage overheads or computational costs. Existing methods on this effort include   efficient architectural designs~\cite{DBLP:journals/corr/HowardZCKWWAA17,DBLP:conf/cvpr/SandlerHZZC18},  pruning~\cite{DBLP:conf/iccv/HeZS17,DBLP:conf/iclr/0022KDSG17}, network quantization~\cite{jmlrHubaraCSEB17,DBLP:conf/icml/LinTA16,DBLP:conf/iclr/ZhouYGXC17,DBLP:conf/iclr/ZhuHMD17}, distillation~\cite{ref10_kd,ref11_feature_kd}, etc. Among these works,  network quantization converts full-precision weights and activations to low-bit discrete values, which can be particularly hardware-friendly. Binary Neural Networks (BNNs)~\cite{corrBengioLC13}, regarded as the extreme case of quantization, resort to  representing networks with $1$-bit values including only $\pm1$. In addition, when weights and activations are both binarized, addition and multiplication operations can be replaced by cheap bit-wise operations. Under this circumstance, BNNs achieve  remarkable compression and acceleration performance. Given the great potentials of BNNs, the community has made numerous efforts to narrow the accuracy gap between full-precision models and BNNs, \emph{e.g.}, introducing scaling factors to lessen the quantization error~\cite{eccvRastegariORF16}, or retaining the kernel information by normalizations~\cite{cvprQinGLSWYS20}. Despite the advances in accuracy, there remains a less-explored direction of further compressing and accelerating BNNs.

A common understanding in the network quantization field is that BNNs are the extreme case enjoying the best compression and acceleration performance, via converting 32-bit floating-point weights/activations into 1-bit ones. However, in this work we challenge this common wisdom. The main goal of this paper is to train a neural network with lower than $1$-bit per weight, which is thus even more compressed than a conventional BNN model. We are motivated by an inspiring observation. Regarding a $3\times 3$ full-precision convolutional kernel, its binarized counterpart belongs to the set which has $|\{\pm1\}^{3\times3}|=512$ different binary kernels. In Figure~\ref{pic:intro} (Left), we visualize the distribution of binary kernels in a binarized ResNet-20~\cite{he2016deep} on CIFAR10~\cite{krizhevsky2009learning}. The distribution indicates that binary kernels in a well-trained BNN model tend to be clustered to a subset at each layer, especially in deep layers where there are more filters.  Such clustering distribution  motivates us that potentially a more compressed model can be attained by learning to identify these important subsets which contain most binary kernels, which is illustrated in Figure~\ref{pic:intro} (Right). For a $3\times3$ convolutional layer with $c_{out}$ output channels, if we choose a subset which consists of $2^\tau$ binary kernels (\emph{e.g.}, $\tau=5$), we will approximately obtain a compression ratio of $\frac{\tau}{9}$ and an acceleration ratio of $\frac{2^\tau}{c_{out}}$ (see Sec.~\ref{subsec:compress} and Sec.~\ref{subsec:accelerate}).

To determine the optimal subsets, we propose a method with two progressive steps. In the first step, we randomly sample layer-specific subsets of binary kernels. To alleviate the potential impact of randomness, in the second step, we refine the subsets by optimization. Models learnt with our full method  are named Sub-bit Neural Networks (SNNs). Figure~\ref{pic:intro} (Right) illustrates the learnt subsets after training an SNN model. With subsets refinement, SNNs tend to learn similar cluster regions as BNNs. \emph{To the best of our knowledge, this is the first method that simultaneously compresses and accelerates BNNs in a quantization pipeline.}

Detailed experiments on image classification verify that our SNNs achieve impressively large compression and acceleration ratios over BNNs while maintaining high accuracies. Besides, the hardware deployment proves that our SNNs using ResNets as test examples bring more than $3\times$ speed up over their BNN counterparts on  ImageNet~\cite{Deng2009ImageNet}.

On the one hand, SNNs introduce a new perspective for the design space and the optimization of binary neural network quantization. Particularly, SNNs uncover and leverage the relationships of clustered binary kernel distributions in different layers of BNN models. On the other hand, SNNs open a new technical direction for compressing and accelerating BNNs while maintaining their hardware-friendly properties, creating potentially valuable opportunities for specialized hardware designs conditioned on the advantages of SNNs.

\section{Related Work}
\label{sec:related_work}
\textbf{Binary neural networks.} Network quantization converts full-precision weights into low-bit counterparts. And in this pipeline, Binary Neural Networks (BNNs)~\cite{corrBengioLC13} are usually treated to be extremely compressed structures. In BNNs, each weight is either $-1$ or $+1$, occupying only $1$-bit which directly leads to a $32\times$ compression ratio compared with a $32$-bit floating-point weight. When activations are binarized together, convolutional operations are equivalent to bit-wise operations. 
Although BNNs are highly efficient and hardware-friendly, one of the main drawbacks is the performance drop. Subsequent works seek various approaches to alleviate this issue. XNOR-Net~\cite{eccvRastegariORF16} reduces the quantization error with channel-wise scaling factors. ABC-Net~\cite{neuripsABCNet} adopts multiple binary bases to approximate the weights and activations to improve the performance. Bi-Real~\cite{ijcvLiuLWYLC20} recommends using short residual connections to reduce the information loss.  IR-Net~\cite{cvprQinGLSWYS20} maximizes the information entropy of binarized parameters apart from minimizing the quantization error, and benefits from the improved information preservation. RBNN~\cite{nipsLinJX00WHL20} learns rotation matrices to reduce the angular bias of the quantization error. Although these works attempt to reduce the performance drop of BNNs, few of them consider the further compression/acceleration for a BNN model. FleXOR~\cite{NeurIPSLeeKKJPY20} converts every flattened binary segment to a shorter segment by encrypting. Although storing encrypted codes can reduce the model size, the complete BNN model needs to be reconstructed before inference and thus FleXOR cannot provide additional benefits in improving the inference speed. 

\textbf{Other efficient designs.} There are other solutions that aim to compress or accelerate a neural network. Structural network pruning methods explicitly prune out filters~\cite{DBLP:conf/iccv/HeZS17,DBLP:conf/iclr/0022KDSG17}. Knowledge distillation methods~\cite{ref10_kd,ref11_feature_kd} can guide the training of a student model with learnt knowledge, such as predictions and features, from a higher-capacity teacher. Some works design lightweight CNN backbones, such as MobileNets \cite{DBLP:journals/corr/HowardZCKWWAA17,DBLP:conf/cvpr/SandlerHZZC18} and ShuffleNets \cite{DBLP:conf/cvpr/ZhangZLS18}. Another pipeline for network efficiency is to adjust the network depths~\cite{DBLP:conf/iclr/CaiGWZH20}, or widths~\cite{DBLP:journals/corr/abs-1903-05134,DBLP:conf/iclr/YuYXYH19}, or resolutions~\cite{wang2020rsnets}. In our work, however, we aim to compress and accelerate a BNN model based on a new binary quantization optimization, and other efficient designs are potentially orthogonal to our method. 

\section{Preliminaries}
\label{sec:pre}
For a CNN model, suppose the weights and activations of its $i$-th layer are denoted by $\rvw^i\in\mathbb{R}^{n^i}$ and $\bm{a}^i\in\mathbb{R}^{m^i}$, where $n^i=c^i_{out}\cdot c^i_{in}\cdot w^i\cdot h^i$ and $m^i=c^i_{out}\cdot W^i\cdot H^i$. Here, $c^i_{out}$ and $c^i_{in}$ are the numbers of output and input channels; $(w^i, h^i)$ and $(W^i, H^i)$ represent the width and height of the weights and activations, respectively. Given the standard convolutional operator $\circledast$, the computation of the $i$-th layer is $\rva^i=\rvw^i\circledast\rva^{i-1}$, with the  bias term omitted for simplicity.

To save the storage and computational costs, BNNs represent weights and/or activations  with 1-bit values ($\pm1$). We denote $\bar{\rvw}^i\in\{\pm1\}^{n^i}$ and $\bar\rva^i\in\{\pm1\}^{m^i}$ as the binary vectors of weights and activations, respectively. The convolution is then reformulated as $\rva^i=\lambda^i\cdot(\bar{\rvw}^i\circ\bar{\rva}^{i-1})$,
where $\lambda^i$ represents channel-wise scaling factors to lessen the quantization error; $\circ$ represents the bitwise operations including XNOR and Bitcount, which can largely improve the computational efficiency compared with the standard convolutional operations.

Among popular methods, the forward pass of weight binarization is simply realized by $\bar{\rvw}^i=\mathrm{sign}(\rvw^i)$, where the element-wise function $\mathrm{sign}(\cdot)$ converts each weight value to $-1$ or $+1$ according to its sign. As the gradient vanishes almost anywhere, the technique named Straight-Through Estimator (STE)~\cite{corrBengioLC13} is widely used to propagate the gradient.

\section{Sub-bit Neural Networks}
\label{sec:methods}
We first introduce our observation which motivates us to propose a two-step approach, \emph{i.e.}, random subsets sampling and subsets refinement. In Sec.~\ref{subsec:compress} and Sec.~\ref{subsec:accelerate}, we discuss that our approach  reduces the model size and increases the inference speed of conventional BNNs. In Sec.~\ref{subsec:deployment}, we  provide a hardware design for the practical deployment.

\subsection{Formulation and Observation}
\label{subsec:insights}

It can be easily proved that binarizing weights with the $\mathrm{sign}(\cdot)$ function is equivalent to clustering weights to the nearest binary vectors. In other words, there is,
\begin{align}
\label{eq:reform}
\bar{\rvw}^i=\mathrm{sign}(\rvw^i)&=\argmin_{\rvb\in\{\pm1\}^{n^i}}\|\rvb-\rvw^i\|_2^2.
\end{align}

In addition, considering the independence of channels, $\bar{\rvw}^i$ can be further derived as the channel-wise concatenation of each binary kernel $\bar{\rvw}^i_c=\argmin_{\rvk\in\mathbb{K}}\|\rvk-\rvw^i_c\|_2^2$, where $\mathbb{K}=\{\pm1\}^{w^i\cdot h^i}$ and $c=1,2,\cdots,c^i_{out}\cdot c^i_{in}$. 

\begin{figure}[t]
\centering
\hskip 1em
\vskip-0.4em
\includegraphics[scale=1.3]{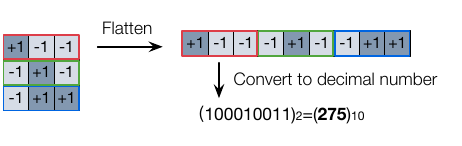}
\vskip-0.4em
\caption{Illustration of assigning an index to a binary kernel. The binary kernel is flattened to a $1$-dimension vector and then is represented with a binary sequence. The assigned index is the corresponding decimal number converted from the  binary sequence.}
\label{pic:convert}
\vskip-0.06in
\end{figure}

Given the definition of $\mathbb{K}$, there are $|\mathbb{K}|=2^{w^i\cdot h^i}$ elements in total, where each element is a binary kernel. We mainly focus on $3\times3$ convolutional kernels as they usually occupy major parameters and computations in modern neural networks, hence here $|\mathbb{K}|=2^9=512$. In BNNs, every single weight occupies $1$-bit, and thus we need $9$-bit to represent a binary kernel. We assign indices (from $1$ to $512$) for these binary kernels, and an example of the assigning process is  depicted in Figure~\ref{pic:convert}. Specifically, we represent the flattened binary kernel with a binary sequence and then convert it to a decimal number. For example, the binary kernel with all $-1$ values is indexed as $1$, and the one with all $+1$ values is indexed as $512$.

\begin{figure}[t]
\centering
\hskip -0.5em
\includegraphics[scale=0.495]{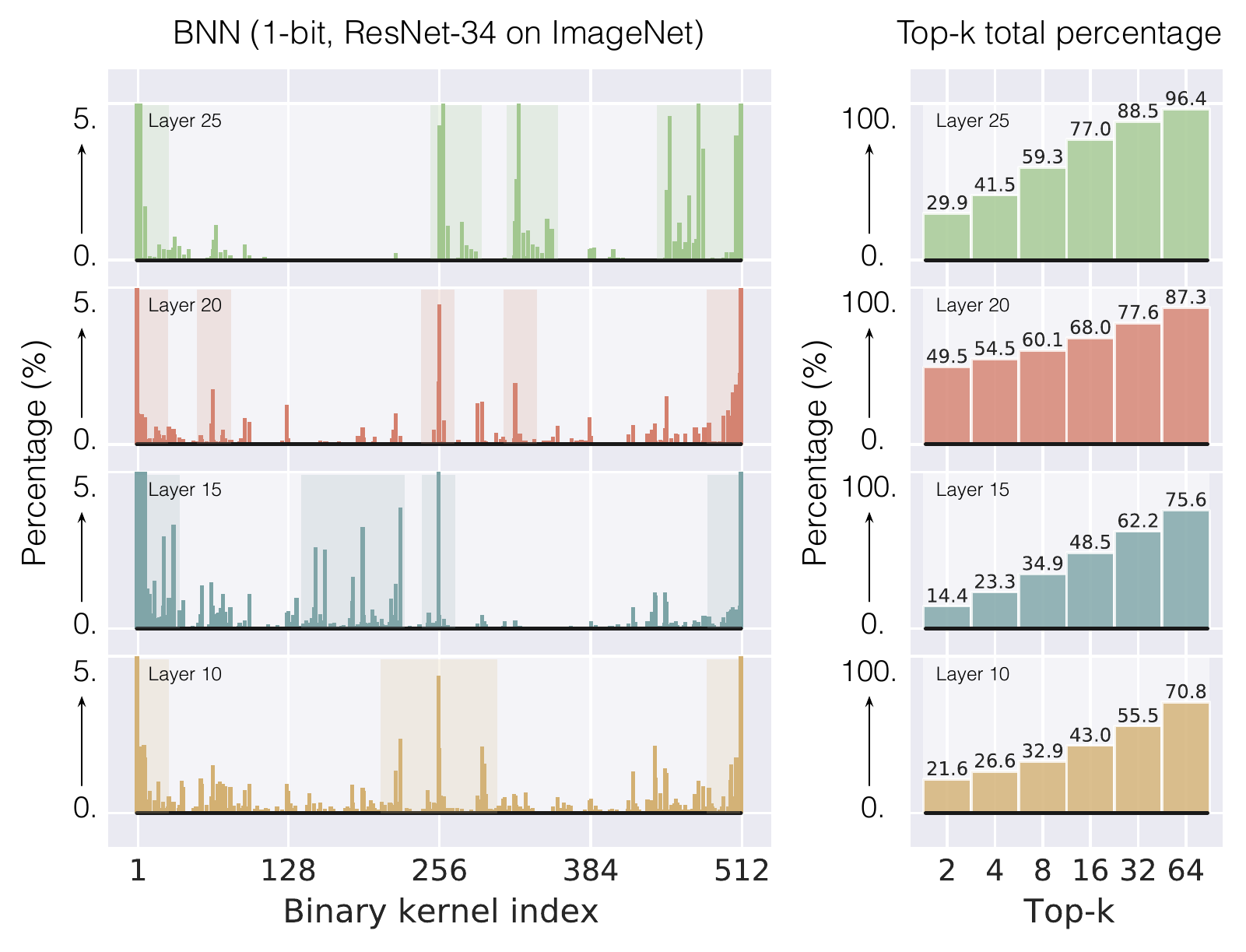}
\caption{\textbf{Left:} Distributions of binary kernels for a standard BNN model which is well-trained on ImageNet using ResNet-34. We do NOT apply any non-linear transformation on the distributions, and only clip the values which are larger than $5\%$. Regions with dense clusters are highlighted with colors. \textbf{Right:} Total percentage of top-k most frequent binary kernels. }
\label{pic:percentage}
\vskip-0.06in
\end{figure}

Here a question arises: how does the  training process of a BNN model tend to distribute the $512$ binary kernels? To figure out this, Figure~\ref{pic:percentage} illustrates the distribution of different binary kernels in a standard BNN model after training, and the total percentage of the top-k most frequent binary kernels in the corresponding layers. Surprisingly, the training process binarizes floating-point kernels to certain clusters/subsets of binary kernels, and given the regions highlighted with colors, the subsets are noticeably different in different layers. The top-k total percentage indicates that, \emph{e.g.}, by only extracting the most frequent $64$ (from the total $512$) binary kernels, the total percentage can surpass $70\%$ in these layers. Besides, in the deep layer with more filters, the clustering property tends to be more obvious. For example, in the 25-th layer,   $96.3\%$ floating-point kernels are binarized to $64$ binary kernels, and  about $60\%$ floating-point kernels are binarized to only $8$ binary kernels.

These findings suggest that a BNN model can be potentially compressed by sampling \emph{layer-specific} subsets of binary kernels and still maintains high performance. Therefore, we design the following two-step method.

\subsection{Random Kernel Subsets Sampling}
\label{subsec:randomsample}
Here we propose a simple yet effective method to generate subsets with binary kernels, which can further reduce the model size and improve inference speed over a BNN model. In contrast with existing binarization methods which convert weights to the full binary set $\mathbb{K}$ (defined in Sec.~\ref{subsec:insights}) without considering kernel structures, we cluster weights in different layers to the different subsets of $\mathbb{K}$, and thus we are able to train a model with lower than $1$ bit per weight. Suppose we want to use $\tau$-bit to represent each kernel (where $1\leq\tau<9$), the size of each subset should be $2^{\tau}$. We set $\tau$ as an integer in our method. Note that we adopt layer-specific subsets instead of sharing the subset throughout the network, in order to maintain the diversity of binary kernels. 

Formally, given the $i$-th layer of a CNN model, we \emph{randomly} sample a subset $\mathbb{P}^i\hskip-0.2em\subset\hskip-0.2em\mathbb{K}$, which satisfies $|\mathbb{P}^i|=2^{\tau}$. We binarize every $\rvw^i_c$ by clustering it to the nearest binary kernel in $\mathbb{P}^i$, and use the standard STE~\cite{corrBengioLC13} method to update $\rvw^i_c$. Under this circumstance, denoting $\mathcal{L}$ as the loss function, the forward and backward passes are computed as,
\begin{flalign}
\label{eq:v1_forward}
&\;\;\;\;\;\;\mathrm{Forward:}\ \bar{\rvw}^i_c=\argmin_{\rvk\in\mathbb{P}^i}\|\rvk-\rvw^i_c\|_2^2,\\
&\;\;\;\;\;\; \mathrm{Backward:}\ \frac{\partial \mathcal{L}}{\partial {\rvw}^i_c}\approx
\begin{cases}
\label{eq:v1_backward}
\frac{\partial \mathcal{L}}{\partial \bar{\rvw}^i_c},& \mathrm{if} \ {\rvw}^i_c \in \left(-1, 1\right),\\
\;\;0,& \mathrm{otherwise}.
\end{cases}&
\end{flalign}

\begin{figure}[t]
\centering
\hskip 0.3em
\includegraphics[scale=0.27]{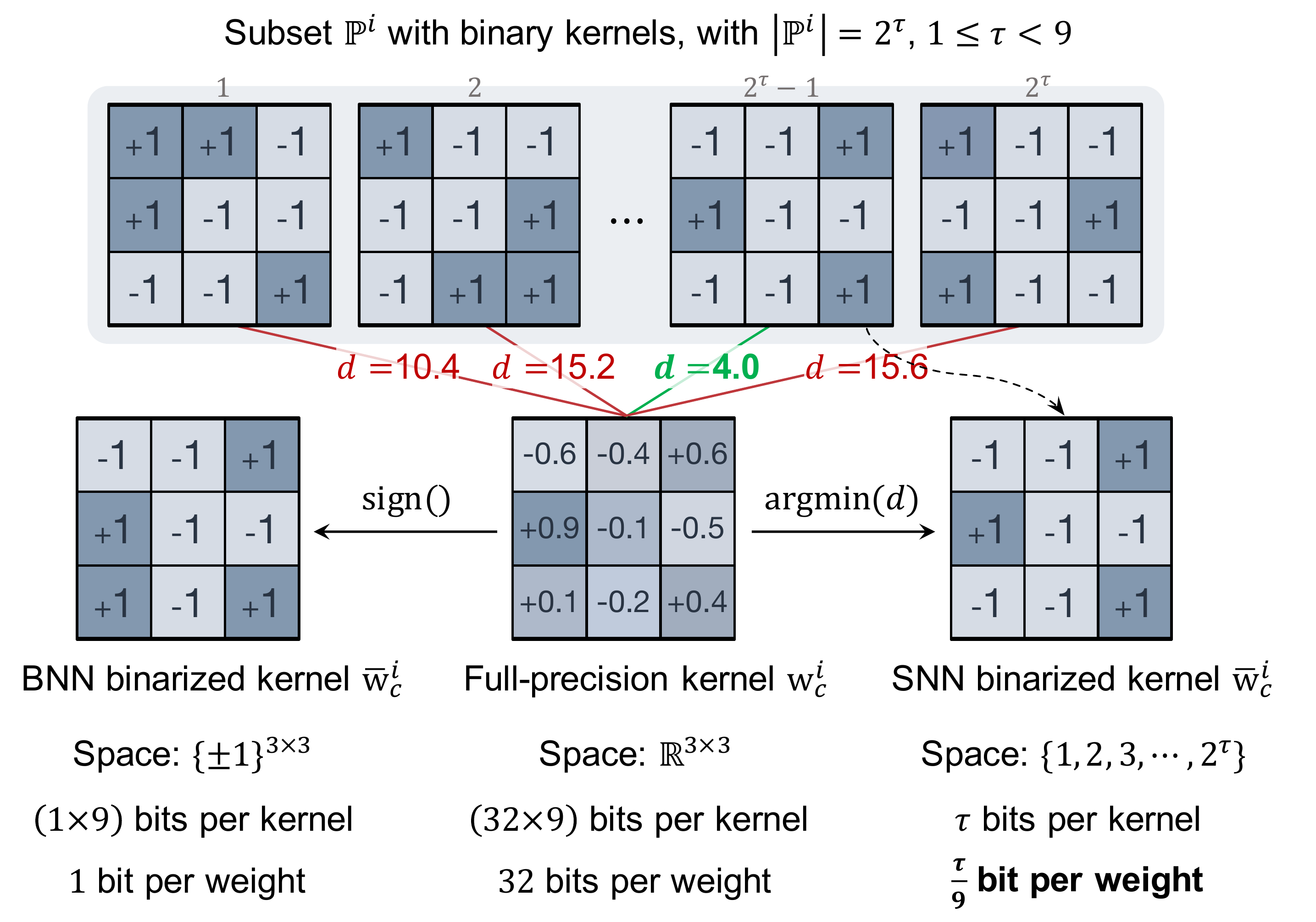}
\caption{Binarization comparison of a standard BNN model and the proposed method (SNN). Here, $d$ denotes $\|\rvk-\rvw^i_c\|_2^2$, as defined in Eq.~(\ref{eq:v1_forward}), where $\rvk\in\mathbb{P}^i$. In our SNN, each kernel corresponds to an index in $\{1,2,3,\cdots,2^\tau\}$, which has the same size with $\{\pm 1\}^\tau$, hence each kernel occupies $\tau$ bits. In average, each weight of SNN only needs $\frac{\tau}{9}$ bit. }
\label{pic:binarize}
\vskip-0.09in
\end{figure}

Figure~\ref{pic:binarize} illustrates why our method can achieve larger compression ratios than a standard BNN model considering the most widely used $3\times 3$ convolutional kernel. Details of the compression are further provided in Sec.~\ref{subsec:compress}. 

This random sampling approach is dubbed the Vanilla version of Sub-bit Neural Networks (Vanilla-SNNs). For this version, the sampling is performed during initialization, and the sampled subsets are fixed throughout the training. Experimental results verify that this method can already demonstrate a good balance between performance and efficiency. In addition, sampling layer-specific subsets  would be less vulnerable to randomness, compared with sharing a global random subset, as will be discussed in Sec.~\ref{subsec:layer_specific}.

\subsection{Kernel Subsets Refinement by Optimization}
\label{subsec:refinement}

Since random sampling cannot guarantee the optimal binary kernels, there is still some room for improvement. To this end, we refine the sampled subsets by optimization.

We represent the subset $\mathbb{P}^i$ with the tensor format $\rvp^i\in\mathbb{R}^{w^i\cdot h^i\cdot|\mathbb{P}^i|}$, which is the concatenation of the kernels. Each single weight in $\rvp^i$ is initialized to $-1.0$ or $+1.0$ according to $\mathbb{P}^i$, and will be updated during training. Since $\rvp^i$ is no longer limited to $\{\pm 1\}$ during optimization, we apply an element-wise function $\mathrm{sign}(\cdot)$ on it. 
Note that the absolute magnitude of each weight of $\rvp^i$ does not affect the forward calculation unless its sign changes. When the input of the $\mathrm{sign}(\cdot)$ function oscillates around 0, the output value rapidly changes between $-1$ and $+1$, which may lead to an unstable training. To tactile this issue, we further introduce a binary tensor $\rvm^i\in\{\pm1\}^{w^i\cdot h^i\cdot|\mathbb{P}^i|}$ to ``memorize'' the sign of each weight in $\rvp^i$. Before training, $\rvm^i$ is initialized to $\mathrm{sign}(\rvp^i)$, and during training, $\rvm^i$ is updated as below,
\begin{flalign}
\label{eq:mask}
\rvm^i=\rvm^i\odot \mathbb{I}_{|\rvp^i|\leq\theta} + \mathrm{sign}(\rvp^i)\odot \mathbb{I}_{|\rvp^i|>\theta},
\end{flalign}
where $\theta$ is the threshold which is a positive and small hyper-parameter for alleviating the rapid sign change and improving the learning stability; $\mathbb{I}$ is the indicator function which obtains a mask tensor belonging to $\{0,1\}^{w^i\cdot h^i\cdot|\mathbb{P}^i|}$; $\odot$ represents the element-wise multiplication. To summarize, $\rvm^i$ follows the sign of $\rvp^i$ only if the magnitude of $\rvp^i$ is larger than $\theta$; otherwise, $\rvm^i$ remains unchanged. $\theta$ is set to $10^{-3}$ in our experiments.

We follow the same expression with Eq. (\ref{eq:v1_backward}) to update $\rvw^i_c$. Similarly, applying the $\mathrm{sign}(\cdot)$ function on $\rvp^i$ in Eq. (\ref{eq:mask}) prevents the gradient propagation, and thus we again adopt the STE technique to update $\rvp^i$ based on the gradient \emph{w.r.t.} $\rvm^i$. The forward pass of binarizing ${\rvw}^i_c$ and the backward pass for updating both ${\rvw}^i_c$ and $\rvp^i$ are computed as,
\begin{flalign}
\label{eq:v2_forward}
&\;\;\;\;\;\;\mathrm{Forward:}\ \bar{\rvw}^i_c=\argmin_{\rvm^i_j}\|\rvm^i_j-\rvw^i_c\|_2^2,\\
&\;\;\;\;\;\; \mathrm{Backward:}\ \text{Eq.}\;(\ref{eq:v1_backward}), \;\;\frac{\partial \mathcal{L}}{\partial \rvp^i}\,\approx\frac{\partial \mathcal{L}}{\partial \rvm^i},&
\end{flalign}
where $j=1,2,\cdots,|\mathbb{P}^i|$; $\rvm^i_j\in\{\pm1\}^{w^i\cdot h^i}$ is the $j$-th binary kernel of $\rvm^i$. Since $\rvm^i$ directly affects $\bar{\rvw}^i_c$ in Eq. (\ref{eq:v2_forward}), the gradient \emph{w.r.t.} $\rvm^i$ could be automatically obtained by accumulating the gradient \emph{w.r.t.} $\bar{\rvw}^i_c$, with $c$ ranging from $1$ to $c^i_{out}\cdot c^i_{in}$. We empirically find that using the same learning rate for updating $\rvw$ and $\rvp$ already yields good results.

With the help of kernel subsets refinement, new binary kernels can be obtained which may be potentially more informative. Given that kernels $\rvp^i_1,\rvp^i_2,\cdots,\rvp^i_{\mathbb{|P}^i|}$ are independently optimized during training, there may appear two repetitive binary kernels in $\rvm^i$ after one certain update, \emph{i.e.}, $\exists\;j_1\ne j_2\;\mathrm{s.t.}\;\rvm^i_{j_1}=\rvm^i_{j_2}$. This issue reduces the number of selectable binary kernels in binarization. In other words, the actual bit-width decreases and may impact expected accuracy. To address the problem, we perform a glance of $\rvm^i$ in every training iteration to check if there are repetitive binary kernels. If repetitive binary kernels are detected in $\rvm^i$, we remove the corresponding kernels in $\rvp^i$ and randomly sample new ones (the same number with the removed ones) from $\mathbb{K}$ to complement $\rvp^i$. By this approach, the aimed bit-width can be  reached during training. These new kernels would together be updated in the later training process.

The learnable approach is dubbed the full version of Sub-bit Neural Networks (SNNs). Forward and backward processes of training SNNs are summarized in Algorithm \ref{algo:training}.

\begin{algorithm}[t]
	\caption{Training: forward and backward processes of Sub-bit Neural Networks (SNNs).} 
	\label{algo:training}
	\small
	\begin{algorithmic}[1]
		\State \textbf{Require}: input data; full-precision weights $\mathbf w$; 
		threshold $\theta$; learning rate $\eta$.
		\State \textbf{for} \text{layer }$i=1\to L \,$ \textbf{do}
		\State \quad Randomly sample a layer-specific subset $\mathbb{P}^i\subset\mathbb{K}$ and there \\ \quad  is $|\mathbb{P}^i|= 2^{\tau}$; Represent $\mathbb{P}^i$ as $\rvp^i\in\mathbb{R}^{w^i\cdot h^i\cdot|\mathbb{P}^i|}$.
		\State \quad Initialize $\rvm^i=\mathrm{sign}(\rvp^i)$.
		\State \textbf{for} \text{step }$t=1\to T\,$ \textbf{do}
		\State \quad{\textbf{Forward propagation:}}
		\State \quad\quad\textbf{for} \text{layer }$i=1\to L\,$ \textbf{do}
		\State \quad\quad\quad Compute $\rvm^i=\rvm^i\odot \mathbb{I}_{|\rvp^i|\leq\theta} + \mathrm{sign}(\rvp^i)\odot \mathbb{I}_{|\rvp^i|>\theta}$.
		\State \quad\quad\quad\textbf{for} \text{channel }$c=1\to c^i_{out}\cdot c^i_{in}$ \textbf{do}
		\State \quad\quad\quad\quad Compute $\bar{\rvw}^i_c =\argmin_{\rvm^i_j}\|\rvm^i_j-\rvw^i_c\|_2^2$.
		\State \quad\quad\quad\quad Compute $\rva^i_c=\lambda^i_c\cdot(\bar{\rvw}^i_c\circ\mathrm{sign}(\rva^{i-1}_c))$ in Sec.~\ref{sec:pre}.
		\State \quad{\textbf{Back propagation:}}
		\State \quad\quad\textbf{for} \text{layer }$i=L\to 1$ \textbf{do}
		\State \quad\quad\quad\textbf{for} \text{channel }$c=1\to c^i_{out}\cdot c^i_{in}$ \textbf{do}
		\State \quad\quad\quad\quad Compute $\frac{\partial \mathcal{L}}{\partial {\rvw}^i_c}$ via Eq. (\ref{eq:v1_backward}).
		\State \quad\quad\quad Compute $\frac{\partial \mathcal{L}}{\partial \rvp^i}\,\approx\frac{\partial \mathcal{L}}{\partial \rvm^i}$.
		
		\State \quad{\textbf{Parameters Update:}}
		
		\State \quad\quad Update $\rvw=\rvw-\eta\frac{\partial{\mathcal{L}}}{\partial{\rvw}}$, $\rvp=\rvp-\eta\frac{\partial{\mathcal{L}}}{\partial{\rvp}}$.
		\State \quad Check repetitive binary kernels in $\rvm^i$ and substitute these corresponding kernels in $\rvp^i$ with random new kernels.
	\end{algorithmic}
\end{algorithm}

\subsection{Compression}
\label{subsec:compress}

\begin{figure}[t]
\centering
\vspace{-0.8em}
\includegraphics[scale=0.57]{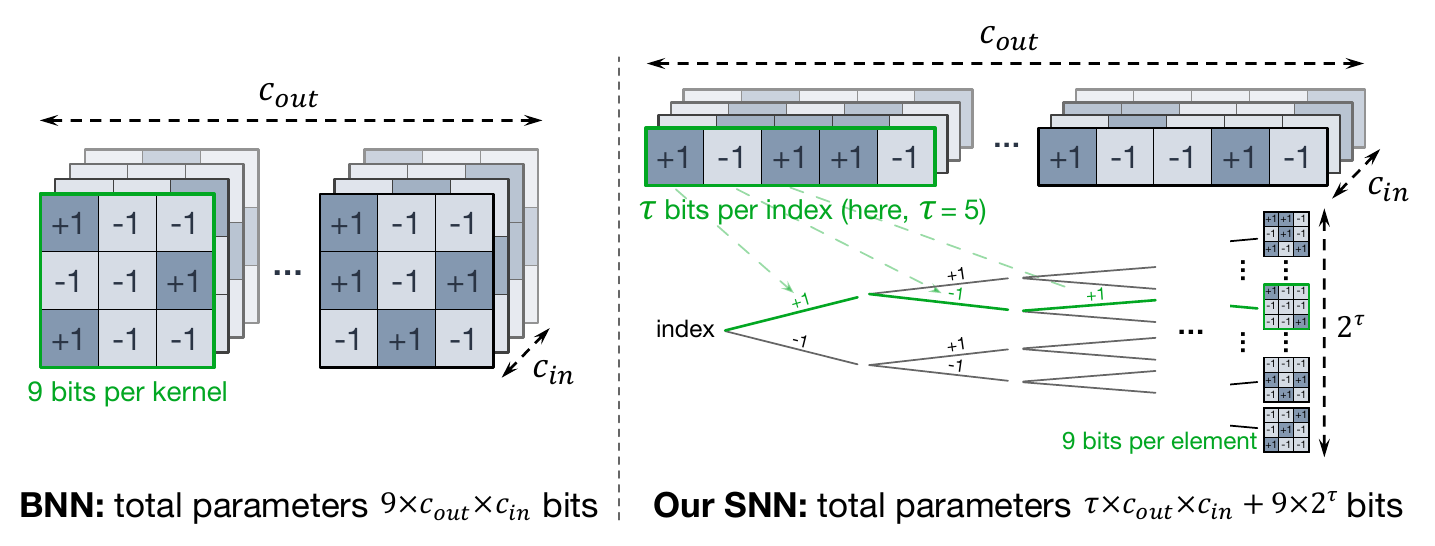}

\caption{Explanation of storing weights and the subset in a layer. }
\label{pic:compression}
\vskip-0.1in
\end{figure}

As preliminarily showed in Figure~\ref{pic:binarize}, in standard BNNs each weight is represented by $1$ bit, and thus each $3\times 3$ binary kernel occupies $9$ bits. To compress BNNs, we adopt an index for each binary kernel, and such index points to one of the $2^\tau$ elements in the subset. Thus the index belongs to $\{1,2,3,\cdots,2^\tau\}$, or equivalently can be represented by a $\tau$-bit $\pm1$ vector. Given $1\leq\tau<9$, every single weight occupies $\frac{\tau}{9}$ bit in average. Here, we further provide Figure~\ref{pic:compression} for a better understanding of the process to store the weights and the subset \emph{in a layer}. According to Figure~\ref{pic:compression}, in SNN, denoting the input and output channel numbers of the layer as $c_{in}$ and $c_{out}$ respectively, weights of this layer occupy $\tau\times c_{in}\times c_{out}$ bits in total. Unlike BNN, we need to additionally store a subset $\mathbb{P}^i$ with $2^\tau$ elements (\emph{i.e.}, $3\times 3$ binary kernels) in  SNN, and such subset occupies $9\times2^\tau$ bits which can be ignored as it is shared per layer. For example, in a $0.56$-bit ($\tau=5$) ResNet-18 or ResNet-34, there are $32$ binary kernels in $\mathbb{P}^i$, which only occupies $0.01\%\hskip-0.2em\sim\hskip-0.2em 0.7\%$ parameters depending on the channel numbers of the layer. As weights of a layer occupy $9\times c_{in}\times c_{out}$ bits in a standard BNN, our SNN achieves a compression ratio of $\frac{\tau}{9}$.

\begin{figure}[t]
\centering
\includegraphics[scale=0.26]{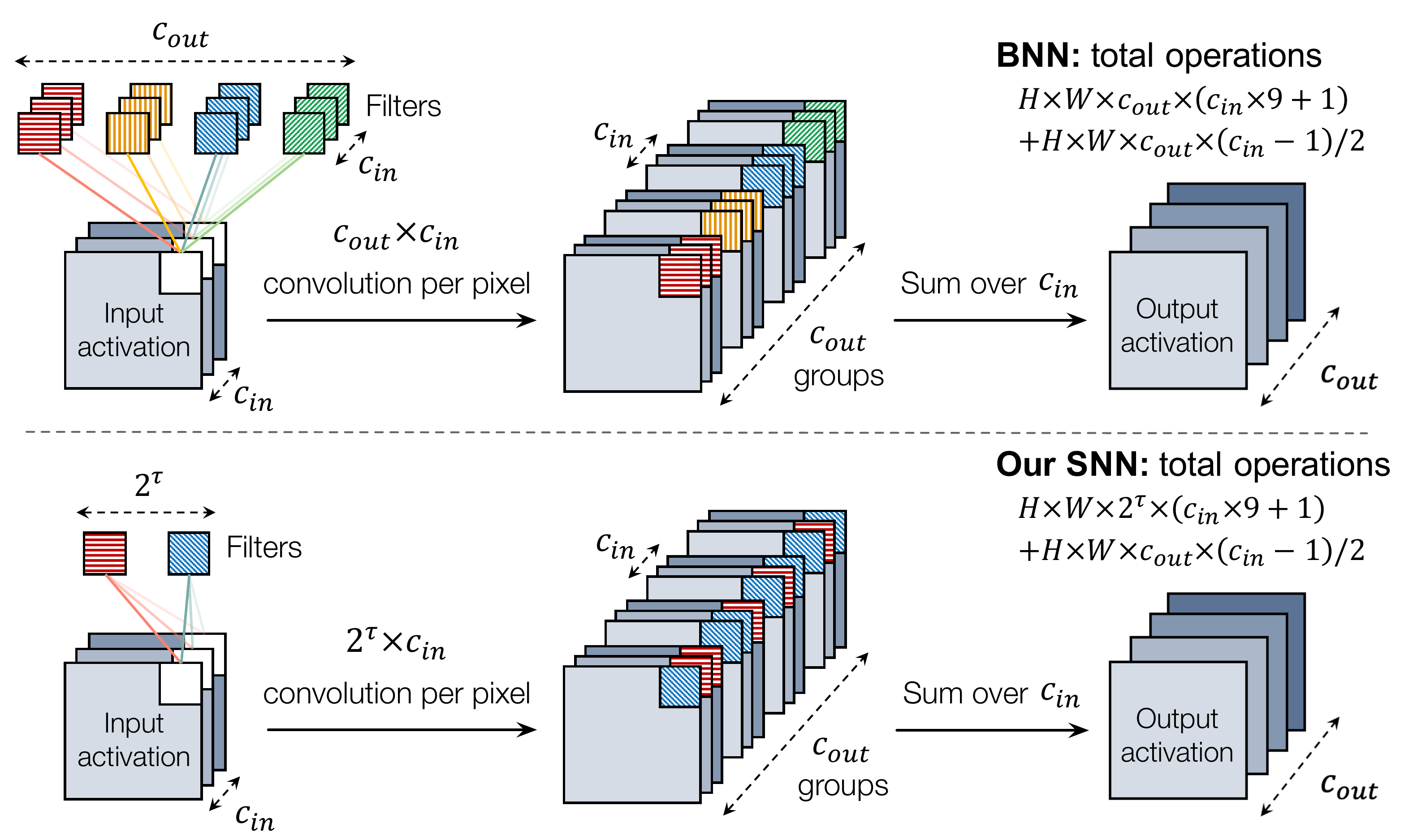}
\caption{Comparison of convolution procedures in a standard BNN  and our SNN. Instead of computing activations for all the $c_{out}$ filters, our method shares the computations, which leads to a larger acceleration as $2^\tau\ll c_{out}$ in most popular backbones such as ResNets. The superscript $i$ is omitted here for simplicity.}
\label{pic:computation}
\vskip-0.08in
\end{figure}

\subsection{Acceleration}
\label{subsec:accelerate}
Apart from compressing the model size, SNNs achieve higher acceleration ratios than standard BNNs.  As the number of different binary kernels is limited to $2^\tau$, convolutional operations can be largely shared. Figure~\ref{pic:computation} depicts the convolution procedures of a standard BNN and our method. The basic difference lies in the convolutional operations, where BNN needs $H \times W \times c^i_{out}\times (c^i_{in}\times 3\times 3+1)$ bit-wise operations for a $3\times3$ convolutional layer, yet SNN reduces this number with a ratio $\frac{2^\tau}{c^i_{out}}$. Both BNN and SNN then perform channel-wise addition along $c^i_{in}$ for $c^i_{out}$ times, and this step occupies much lower computational burdens compared with the previous step (convolution). Note that SNN needs additional indexing operations to reuse the calculated activations, and such indexing costs can be largely saved by designing the data flow as will be introduced in Sec.~\ref{subsec:deployment}.

\subsection{Practical Deployment}
\label{subsec:deployment}

\begin{figure}[t]
\centering\vskip-0.3em
\hskip -0.1em
\includegraphics[scale=0.35]{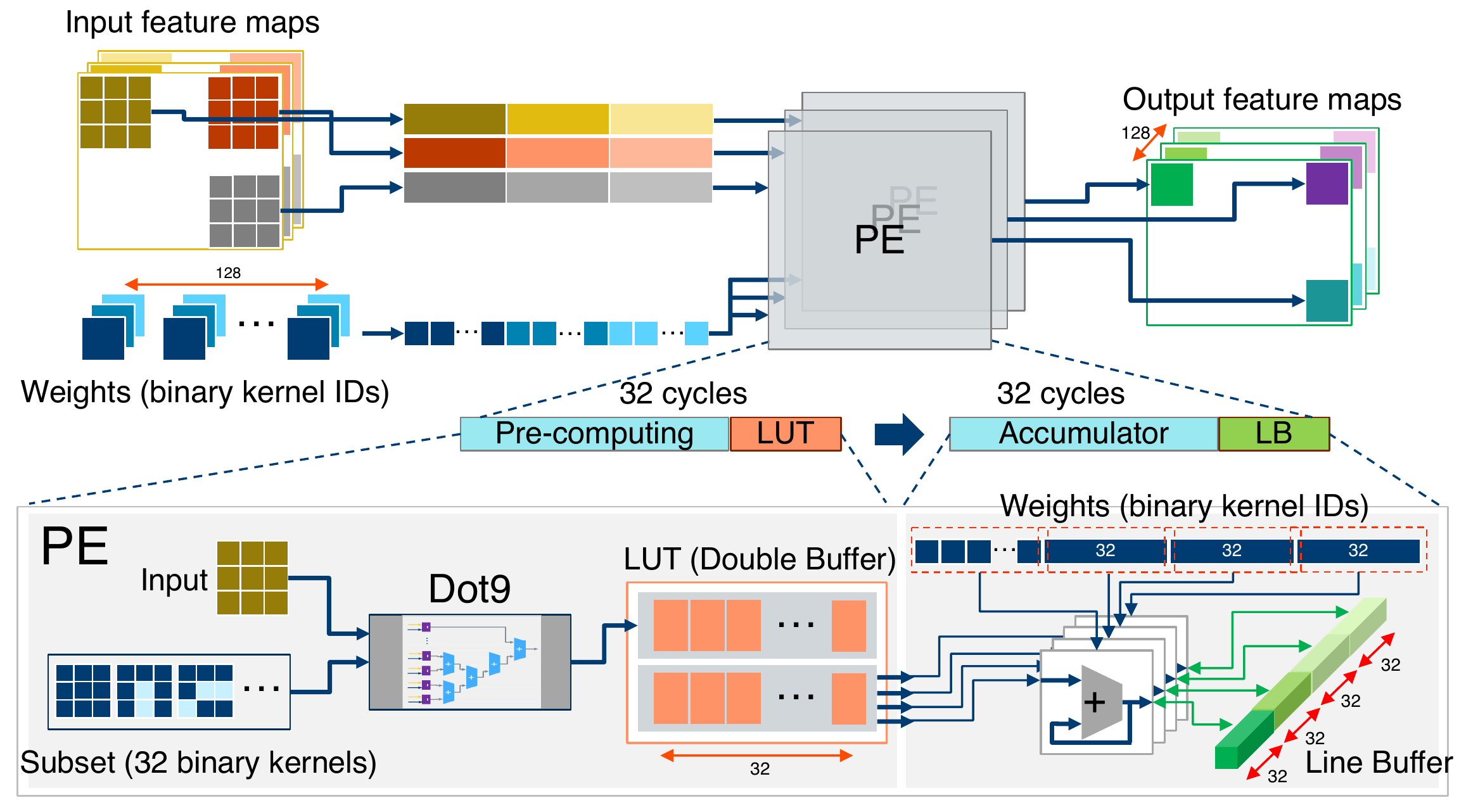}
\vskip0.3em
\caption{Hardware design for the deployment with a $0.56$-bit SNN for example, where the subset of each layer contains $32$ binary kernels. The pre-computing unit needs $32$ cycles to traverse all binary kernels. The accumulator unit has $4$ parallel accumulators, which accumulate $4$ groups of data in one cycle and write the results back to the corresponding positions of LB simultaneously. To match the number of cycles of the pipeline stage, the width of LB could be set to $128$, so that it also needs $32$ cycles to complete the process. 
}
\label{pic:simulation}
\vskip-0.1in
\end{figure}

We have discussed that our SNNs are theoretically more efficient than standard BNNs. To verify the runtime acceleration in practical deployment, we present a hardware accelerator architecture for SNNs in Figure~\ref{pic:simulation}.
Specially, we divide the Processing Engine (PE) into two pipeline units. First, the pre-computing unit fetches binary kernels serially by kernel IDs and convolutes them with input activations. The result is immediately stored in the Lookup Table (LUT) within the current cycle.  Second, the accumulator unit reads kernel IDs and fetches the data from LUT with kernel IDs as addresses. The data from LUT will be accumulated with the data stored in a Line Buffer (LB), and then will be written back to LB. Once all input activations along input channels are processed, data in LB will be stored into a dedicated output buffer and cleared for the next round. Such architecture design is particularly FPGA friendly: as FPGA itself is a LUT-based structure, the cost of implementing the distributed lookup tables is very small. Both units support parallel computing in different PEs. Two units follow a data pipeline and are parallelly computed within the same cycles.

Theoretically, each binary kernel in the subset $\mathbb{P}^i$ is pre-computed per channel and per pixel of the $c_{in}\times H^i\times W^i$ input activation, and thus there are $2^\tau\times c_{in}\times H^i\times W^i$ pre-computed results need storing in LUT. However, by well designing the computation flow, we can largely reduce the LUT size and thus decrease the lookup time costs. As depicted in Figure~\ref{pic:simulation}, the $c_{in}\times H^i\times W^i$ input activation is split into $1\times 3\times3$ activation slices which are overlapped (given the convolutional stride 1) and are computed in parallel. At each time, all pre-computed results in LUT only correspond to one same activation slice. Therefore the practical LUT size is no longer $2^\tau\times c_{in}\times H^i\times W^i$ but is reduced to only $2^\tau$. Such a small LUT size leads to very low latency for the lookup process, \emph{e.g.}, less than 0.5ns for our $0.56$-bit model which can be easily implemented in the current clock cycle. Thus the lookup process is well integrated into the deployment and may not affect the number of cycles.

\section{Experiments}
We conduct comprehensive experiments to evaluate the effectiveness and efficiency of our method. All our experiments are implemented with the PyTorch~\cite{neuripsPaszkeGMLBCKLGA19} library.

\subsection{Basic Results}
We choose CIFAR10~\cite{krizhevsky2009learning} and ImageNet~\cite{Deng2009ImageNet} datasets for experiments. Following IR-Net~\cite{cvprQinGLSWYS20}, the Bi-Real~\cite{ijcvLiuLWYLC20} technique is adopted when activations are also binarized. Following the common paradigm~\cite{ijcvLiuLWYLC20,cvprQinGLSWYS20,eccvRastegariORF16}, we keep the first and last layers to be full-precision and binarize the others.

\begin{table}[t]
	\centering\vskip -0.05in
	\caption{Image classification results on  CIFAR10 based on the single-crop testing with $32 \times 32$ crop size. We follow the training details and techniques in IR-Net~\cite{cvprQinGLSWYS20}, which could be the reference for our $1$-bit baselines. Each result of our method is the average of three runs.  Numbers highlighted in green are reduction ratios over BNN counterparts. $^*$ indicates our implemented results.
	} \vskip 0.05in
	\tablestyle{4pt}{1}
	\resizebox{1\linewidth}{!}{
		\begin{tabular}{c|c|c|c|c}
			\label{tab:cifar10}
			\fontsize{8pt}{1em}\selectfont \multirow{2}*{Method} 
			& \fontsize{8pt}{1em}\selectfont Bit-width
			& \fontsize{8pt}{1em}\selectfont \#Params
			& \fontsize{8pt}{1em}\selectfont Bit-OPs
			& \fontsize{8pt}{1em}\selectfont Top-1 Acc.\\
			&\fontsize{8pt}{1em}\selectfont (W$/$A)
			&\fontsize{8pt}{1em}\selectfont  (Mbit)
			& \fontsize{8pt}{1em}\selectfont (G)
			& \fontsize{8pt}{1em}\selectfont (\%)
			\\
			\shline
			\multicolumn{5}{c}{\fontsize{8pt}{1em}\selectfont {ResNet-18}}\\
			\cdashline{1-5}[1pt/1pt]
			Full precision &32$/$32&351.54 &35.03  & 93.0 \\
			\cdashline{1-5}[1pt/1pt]
			RAD~\cite{Regularize-act-distribution} & 1$/$1 &10.99 & 0.547& 90.5 \\
			IR-Net~\cite{cvprQinGLSWYS20} &1$/$1&10.99  & 0.547 & 91.5 \\
			\rowcolor{cyan!7}
			Vanilla-SNN $\vert$ SNN & 0.67$/$1 &7.324\cgaphltimes{1.5}{$\times$}&0.289\cgaphltimes{1.9}{$\times$}& 89.7 $\vert$ 91.0 \\
			\rowcolor{cyan!7}
			Vanilla-SNN $\vert$ SNN & 0.56$/$1 &6.103\cgaphltimes{1.8}{$\times$}& 0.164\cgaphltimes{3.3}{$\times$}& 89.3 $\vert$ 90.6 \\
			\rowcolor{cyan!7}
			Vanilla-SNN $\vert$ SNN & 0.44$/$1 &4.882\cgaphltimes{2.3}{$\times$}&0.097\cgaphltimes{5.6}{$\times$}& 88.3 $\vert$ 90.1  \\
			
			\cdashline{1-5}[1pt/1pt]
			IR-Net$^*$~\cite{cvprQinGLSWYS20}  & 1$/$32 & 10.99& 17.52& 92.9 \\
			\rowcolor{cyan!7}
			Vanilla-SNN $\vert$ SNN & 0.67$/$32 & 7.324\cgaphltimes{1.5}{$\times$}& 9.236\cgaphltimes{1.9}{$\times$}& 92.4 $\vert$ 92.7 \\
			\rowcolor{cyan!7}
			Vanilla-SNN $\vert$ SNN & 0.56$/$32 &6.103\cgaphltimes{1.8}{$\times$}&5.239\cgaphltimes{3.3}{$\times$}& 92.0 $\vert$ 92.3 \\
			\rowcolor{cyan!7}
			Vanilla-SNN $\vert$ SNN & 0.44$/$32&4.882\cgaphltimes{2.3}{$\times$}&3.106\cgaphltimes{5.6}{$\times$}& 91.6 $\vert$ 91.9  \\
			\hline

			\multicolumn{5}{c}{\fontsize{8pt}{1em}\selectfont {ResNet-20}}\\
			\cdashline{1-5}[1pt/1pt]
			Full precision  &32$/$32& 8.54&  2.567& 91.7 \\
			\cdashline{1-5}[1pt/1pt]
			DoReFa~\cite{corrdorefa} & 1$/$1 &0.267 &0.040& 79.3 \\
			IR-Net~\cite{cvprQinGLSWYS20} &1$/$1&0.267  &  0.040& 86.5 \\
			\rowcolor{cyan!7}
			\rowcolor{cyan!7}
			Vanilla-SNN $\vert$ SNN & 0.67$/$1 &0.178\cgaphltimes{1.5}{$\times$}& 0.040& 83.9 $\vert$ 85.1 \\
			\rowcolor{cyan!7}
			Vanilla-SNN $\vert$ SNN & 0.56$/$1 &0.148\cgaphltimes{1.8}{$\times$}&0.034\cgaphltimes{1.2}{$\times$}& 82.7 $\vert$ 84.0 \\
			\rowcolor{cyan!7}
			Vanilla-SNN $\vert$ SNN & 0.44$/$1 &0.119\cgaphltimes{2.3}{$\times$}& 0.025\cgaphltimes{1.6}{$\times$}& 82.0 $\vert$ 82.5  \\
			\cdashline{1-5}[1pt/1pt]
			DoReFa~\cite{corrdorefa} & 1$/$32 &0.267 &1.283 & 90.0 \\
			LQ-Net~\cite{eccvLQNet} &1$/$32&0.267  & 1.283 & 90.1 \\
			IR-Net~\cite{cvprQinGLSWYS20} &1$/$32& 0.267 & 1.283 & 90.8 \\
			\rowcolor{cyan!7}
			Vanilla-SNN $\vert$ SNN & 0.67$/$32 &0.178\cgaphltimes{1.5}{$\times$}& 1.283& 88.7 $\vert$ 90.0 \\
			\rowcolor{cyan!7}
			Vanilla-SNN $\vert$ SNN & 0.56$/$32 &0.148\cgaphltimes{1.8}{$\times$}& 1.099\cgaphltimes{1.2}{$\times$}& 87.8 $\vert$ 88.9 \\
			\rowcolor{cyan!7}
			Vanilla-SNN $\vert$ SNN & 0.44$/$32 &0.119\cgaphltimes{2.3}{$\times$}& 0.822\cgaphltimes{1.6}{$\times$}& 87.1 $\vert$ 87.6  \\
			\hline
			\multicolumn{5}{c}{\fontsize{8pt}{1em}\selectfont {VGG-small}}\\
			\cdashline{1-5}[1pt/1pt]
			Full precision &32$/$32&146.24 &  38.66& 92.5 \\
			\cdashline{1-5}[1pt/1pt]
			LAB~\cite{Loss-Aware-BNN} & 1$/$1 &4.571 &0.603 & 87.7 \\
			XNOR~\cite{eccvRastegariORF16} &1$/$1& 4.571 & 0.603 & 89.8 \\
			BNN~\cite{nipsHubaraCSEB16} &1$/$1&4.571  & 0.603 & 89.9 \\
			RAD~\cite{Regularize-act-distribution} &1$/$1&4.571  & 0.603 & 90.0 \\
			IR-Net$^*$~\cite{cvprQinGLSWYS20} &1$/$1& 4.571 & 0.603 & 91.3 \\
			\rowcolor{cyan!7}
			Vanilla-SNN $\vert$ SNN & 0.67$/$1 &3.047\cgaphltimes{1.5}{$\times$}& 0.194\cgaphltimes{3.1}{$\times$}& 90.3 $\vert$ 91.0 \\
			\rowcolor{cyan!7}
			Vanilla-SNN $\vert$ SNN & 0.56$/$1 &2.540\cgaphltimes{1.8}{$\times$}&  0.113\cgaphltimes{5.3}{$\times$}& 89.8 $\vert$ 90.6 \\
			\rowcolor{cyan!7}
			Vanilla-SNN $\vert$ SNN & 0.44$/$1 &2.032\cgaphltimes{2.3}{$\times$}& 0.074\cgaphltimes{8.1}{$\times$}& 89.2 $\vert$ 90.0  \\
			\cdashline{1-5}[1pt/1pt]
			IR-Net$^*$~\cite{cvprQinGLSWYS20} &1$/$32& 4.571 & 19.30 & 92.5 \\
			\rowcolor{cyan!7}
			Vanilla-SNN $\vert$ SNN & 0.67$/$32 &3.047\cgaphltimes{1.5}{$\times$}& 6.208\cgaphltimes{3.1}{$\times$}& 92.0 $\vert$ 92.4 \\
			\rowcolor{cyan!7}
			Vanilla-SNN $\vert$ SNN & 0.56$/$32 &2.540\cgaphltimes{1.8}{$\times$}&  3.616\cgaphltimes{5.3}{$\times$}& 91.7 $\vert$ 92.1 \\
			\rowcolor{cyan!7}
			Vanilla-SNN $\vert$ SNN & 0.44$/$32 &2.032\cgaphltimes{2.3}{$\times$}& 2.368\cgaphltimes{8.1}{$\times$}& 91.3 $\vert$ 91.9  \\

		\end{tabular}
	}
\vskip -0.1in
\end{table}

\begin{table}[t]
	\centering\vskip -0.05in
	\caption{Image classification results on ImageNet based on the single-crop testing  with  $224 \times 224$ crop size. We follow the training details and techniques in IR-Net~\cite{cvprQinGLSWYS20}, which could be the reference for our $1$-bit baselines. Numbers highlighted in green are reduction ratios over BNN counterparts.
	} \vskip 0.05in
	\tablestyle{6pt}{1.035}
	\resizebox{1\linewidth}{!}{
		\begin{tabular}{c|c|c|c|c}
			\label{tab:imagenet}
			\fontsize{8pt}{1em}\selectfont \multirow{2}*{Method} 
			& \fontsize{8pt}{1em}\selectfont Bit-width
			& \fontsize{8pt}{1em}\selectfont \#Params
			& \fontsize{8pt}{1em}\selectfont Bit-OPs
			& \fontsize{8pt}{1em}\selectfont Top-1 Acc.\\
			&\fontsize{8pt}{1em}\selectfont (W$/$A)
			&\fontsize{8pt}{1em}\selectfont  (Mbit)
			& \fontsize{8pt}{1em}\selectfont (G)
			& \fontsize{8pt}{1em}\selectfont (\%)
			\\
			\shline
			\multicolumn{5}{c}{\fontsize{8pt}{1em}\selectfont {ResNet-18}}\\
			\cdashline{1-5}[1pt/1pt]
			Full precision &32$/$32& 351.54& 107.28 & 69.6 \\
			\cdashline{1-5}[1pt/1pt]
			XNOR~\cite{eccvRastegariORF16} & 1$/$1 & 10.99& 1.677& 51.2 \\
			BNN+~\cite{nipsHubaraCSEB16} & 1$/$1 & 10.99& 1.677& 53.0 \\
			Bi-Real~\cite{ijcvLiuLWYLC20} & 1$/$1 & 10.99& 1.677& 56.4 \\
			XNOR++~\cite{bmvcBulatT19} & 1$/$1 &10.99 &1.677 & 57.1 \\
			IR-Net~\cite{cvprQinGLSWYS20} & 1$/$1 & 10.99&1.677 & 58.1 \\
			\rowcolor{cyan!7}
			Vanilla-SNN $\vert$ SNN & 0.67$/$1 &7.324\cgaphltimes{1.5}{$\times$}& 0.883\cgaphltimes{1.9}{$\times$}& 55.7 $\vert$ 56.3 \\
			\rowcolor{cyan!7}
			Vanilla-SNN $\vert$ SNN & 0.56$/$1 &6.103\cgaphltimes{1.8}{$\times$}& 0.501\cgaphltimes{3.3}{$\times$}& 54.6 $\vert$ 55.1 \\
			\rowcolor{cyan!7}
			Vanilla-SNN $\vert$ SNN & 0.44$/$1 &4.882\cgaphltimes{2.3}{$\times$}& 0.297\cgaphltimes{5.6}{$\times$}& 52.5 $\vert$ 53.0  \\
			\cdashline{1-5}[1pt/1pt]
			BWN~\cite{eccvRastegariORF16} & 1$/$32 & 10.99&53.64& 60.8 \\
			HWGQ~\cite{iccvhwgq} & 1$/$32 &10.99&53.64& 61.3 \\
			BWHN~\cite{DBLP:journals/corr/abs-1802-02733} & 1$/$32 & 10.99&53.64& 64.3 \\
			IR-Net~\cite{cvprQinGLSWYS20} &1$/$32&  10.99& 53.64& 66.5 \\
			FleXOR~\cite{NeurIPSLeeKKJPY20} &0.80$/$32& 8.788\cgaphltimes{1.3}{$\times$}& 53.64& 63.8 \\
			FleXOR~\cite{NeurIPSLeeKKJPY20} &0.60$/$32& 6.591\cgaphltimes{1.7}{$\times$}& 53.64& 62.0 \\
			\rowcolor{cyan!7}
			Vanilla-SNN $\vert$ SNN & 0.67$/$32 &7.324\cgaphltimes{1.5}{$\times$}& 28.26\cgaphltimes{1.9}{$\times$}& 63.7 $\vert$ 64.7 \\
			\rowcolor{cyan!7}
			Vanilla-SNN $\vert$ SNN & 0.56$/$32 & 6.103\cgaphltimes{1.8}{$\times$}& 16.03\cgaphltimes{3.3}{$\times$}& 62.8 $\vert$ 63.4 \\
			\rowcolor{cyan!7}
			Vanilla-SNN $\vert$ SNN & 0.44$/$32 & 4.882\cgaphltimes{2.3}{$\times$}&9.504\cgaphltimes{5.6}{$\times$}& 60.1 $\vert$ 60.9  \\
			\hline
			\multicolumn{5}{c}{\fontsize{8pt}{1em}\selectfont {ResNet-34}}\\
			\cdashline{1-5}[1pt/1pt]
			Full precision &32$/$32&674.88 & 225.66 & 73.3 \\
			\cdashline{1-5}[1pt/1pt]
			Bi-Real~\cite{ijcvLiuLWYLC20} &1$/$1&21.09& 3.526& 62.2 \\
			IR-Net~\cite{cvprQinGLSWYS20} &1$/$1& 21.09& 3.526& 62.9 \\
			\rowcolor{cyan!7}
			Vanilla-SNN $\vert$ SNN & 0.67$/$1 &14.06\cgaphltimes{1.5}{$\times$}&1.696\cgaphltimes{2.1}{$\times$}& 60.6 $\vert$ 61.4 \\
			\rowcolor{cyan!7}
			Vanilla-SNN $\vert$ SNN & 0.56$/$1 &11.71\cgaphltimes{1.8}{$\times$}&0.965\cgaphltimes{3.7}{$\times$}& 59.5 $\vert$ 60.2 \\
			\rowcolor{cyan!7}
			Vanilla-SNN $\vert$ SNN & 0.44$/$1 &9.372\cgaphltimes{2.3}{$\times$}& 0.581\cgaphltimes{6.1}{$\times$}& 58.1 $\vert$  58.6\\
			\cdashline{1-5}[1pt/1pt]
			IR-Net~\cite{cvprQinGLSWYS20} &1$/$32&21.09 &112.83& 70.4 \\
			\rowcolor{cyan!7}
			Vanilla-SNN $\vert$ SNN & 0.67$/$32 &14.06\cgaphltimes{1.5}{$\times$}&54.27\cgaphltimes{2.1}{$\times$}&  67.5 $\vert$ 68.0 \\
			\rowcolor{cyan!7}
			Vanilla-SNN $\vert$ SNN & 0.56$/$32 & 11.71\cgaphltimes{1.8}{$\times$}& 30.88\cgaphltimes{3.7}{$\times$}&  66.3 $\vert$ 66.9 \\
			\rowcolor{cyan!7}
			Vanilla-SNN $\vert$ SNN & 0.44$/$32 &9.372\cgaphltimes{2.3}{$\times$}& 18.59\cgaphltimes{6.1}{$\times$}& 64.5 $\vert$  65.1\\

		\end{tabular}
	}
\vskip-0.135in
\end{table}

\textbf{CIFAR10.} Following popular settings in state-of-the-art BNNs, we adopt ResNet-20~\cite{he2016deep}, ResNet-18, and VGG-small~\cite{DBLP:journals/corr/SimonyanZ14a} to evaluate our method on  CIFAR10. Results including our Vanilla-SNNs, SNNs, and other existing methods are provided in Table~\ref{tab:cifar10}. For  our method, we use three typical bit-widths $0.67$-bit, $0.56$-bit, and $0.44$-bit, which are obtained by setting $\tau$ to $6$, $5$, and $4$ respectively. Bit-wise parameters (Params) and bit-wise operations (Bit-OPs) are compared to highlight the superiority of our method\footnote{For comparison, in Table~\ref{tab:cifar10} and Table~\ref{tab:imagenet}, parameters and bit-wise operations for all methods are collected excluding the first and last layers as they are not binarized following the common paradigm.}.  Our method presents increasing compression and acceleration rates as the bit-width decreases, with slight and stable performance variations. The acceleration rate is highly correlated to the number of channels in the architecture. For example, by compressing BNN to $0.56$-bit in VGG-small, SNN has almost no accuracy drop yet reduces the model size and Bit-OPs by $1.80\times$ and $5.34\times$ respectively. 

\textbf{ImageNet.} We also conduct experiments on ImageNet, one of the most challenging benchmarks for visual recognition. In Table~\ref{tab:imagenet}, we report results based on ResNet-18 and ResNet-34 with bit-width settings $0.67$-bit, $0.56$-bit, and $0.44$-bit to evaluate our method. Compared with state-of-the-art BNN methods, in general, SNNs with $0.67$-bit, $0.56$-bit, and $0.44$-bit have around $1.5\%$, $3.0\%$, $4.5\%$ accuracy drops  respectively, yet saving the parameters by  $1.50\times$, $1.80\times$, $2.25\times$, and saving Bit-OPs by $1.90\hskip-0.02in\sim\hskip-0.02in 2.08\times$, $3.35\hskip-0.02in\sim\hskip-0.02in 3.65\times$, $5.65\hskip-0.02in\sim\hskip-0.02in 6.07\times$. By reducing the bit-width from $1$-bit to $0.44$-bit, the number of selectable binary kernels for each layer is significantly reduced from $512$ to $16$. Considering this, the accuracy drop is moderate. As introduced in Sec.~\ref{sec:related_work}, FleXOR~\cite{NeurIPSLeeKKJPY20} reduces parameters by encryption and relies on decryption  before inference, leading no further benefits to the running speed. Table~\ref{tab:imagenet} shows that our SNNs outperform FleXORs in two aspects: 1) SNNs achieve better accuracies with lower bit-width settings, \emph{e.g.}, SNNs with $0.56/32$-bit achieving $63.4\%$ vs. FleXORs with $0.6/32$-bit achieving $62.0\%$; 2) SNNs largely reduce inference Bit-OPs. In addition, SNNs obtain better performance than the corresponding Vanilla-SNNs, \emph{e.g.}, from $62.8\%$ to $63.4\%$ for $0.56/32$-bit ResNet-18, which verifies the advantage of subsets refinement proposed in Sec.~\ref{subsec:refinement}.

\subsection{Ablation Studies}
\textbf{Layer-specific or layer-shared kernel subsets.}\label{subsec:layer_specific} The observation in Sec.~\ref{subsec:insights} suggests that binary kernels tend to cluster to different subsets in different layers, and thus the random sampling process in Sec.~\ref{sec:methods} privatizes subsets for different layers. Using layer-specific subsets also alleviates the instability brought by randomness. Comparison results in Table~\ref{tab:specific}  indicate that: 1) Settings with layer-shared subsets present unstable results especially at a low bit-width such as $0.11$ or $0.33$, while using layer-specific subsets is more effective and stable. 2) Kernel subsets refinement, proposed in Sec.~\ref{subsec:refinement}, can  improve the results for both the layer-specific case as well as the layer-shared  case.

\begin{figure}[t]
\centering
\hskip -0.1em
\includegraphics[scale=0.4]{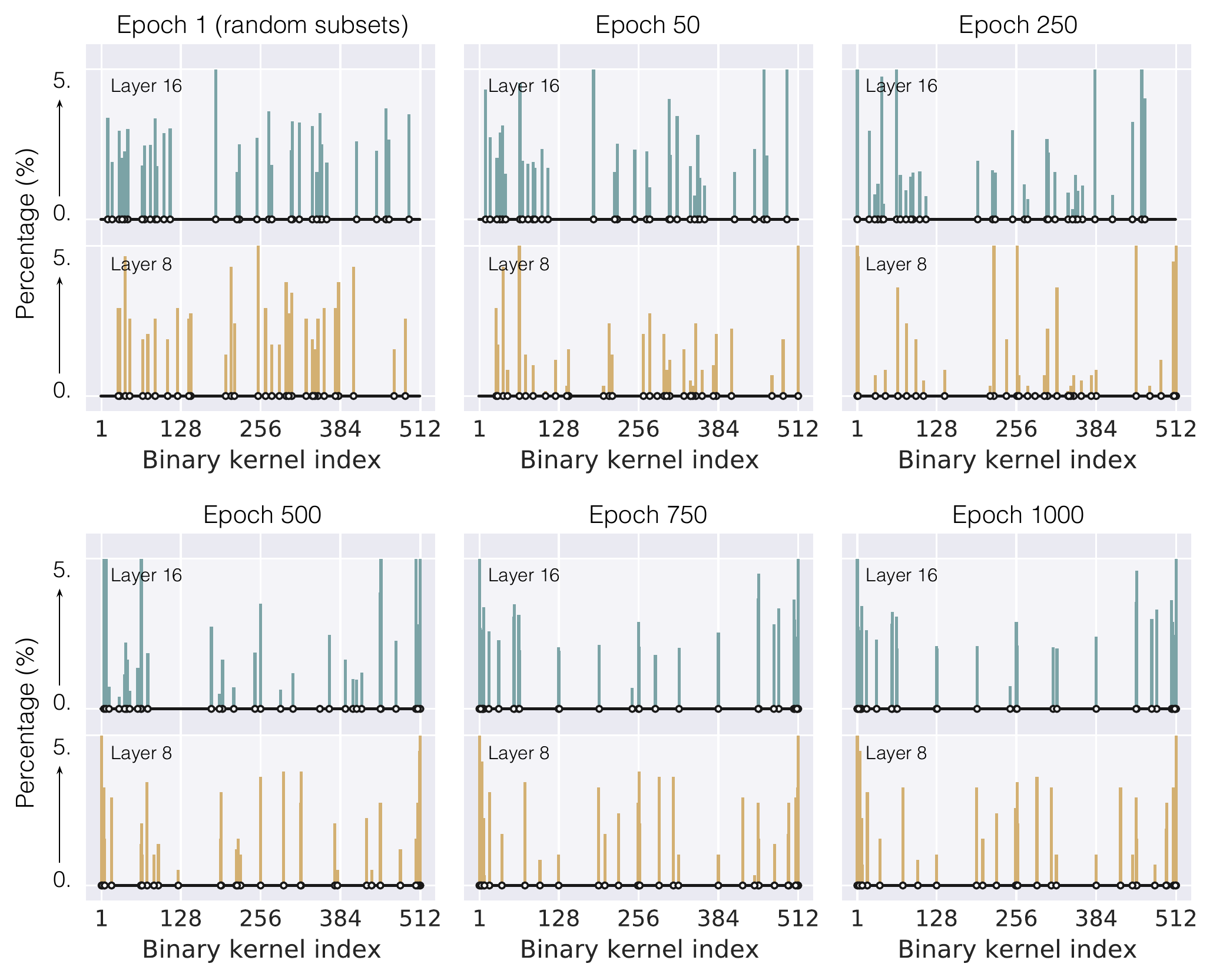}
\caption{Visualization of how binary kernel subsets change during the training process of a $0.56$-bit SNN. The experiment is performed on CIFAR10 with ResNet-20. }
\label{pic:epoch}
\end{figure}

\textbf{Learning procedure of refinement.} Although  experiments have verified the validity of subsets refinement, we would like to figure out why it works. Figure~\ref{pic:epoch} illustrates the changes of subsets during training (with subsets refinement). We observe that binary kernels seem to be irregular at the first epoch since they are initially randomly sampled. With the help of subsets refinement, distributions of binary kernels tend to be regular and symmetric in later epochs.

\begin{figure}[t]
\centering
\hskip -0.3em
\includegraphics[scale=0.37]{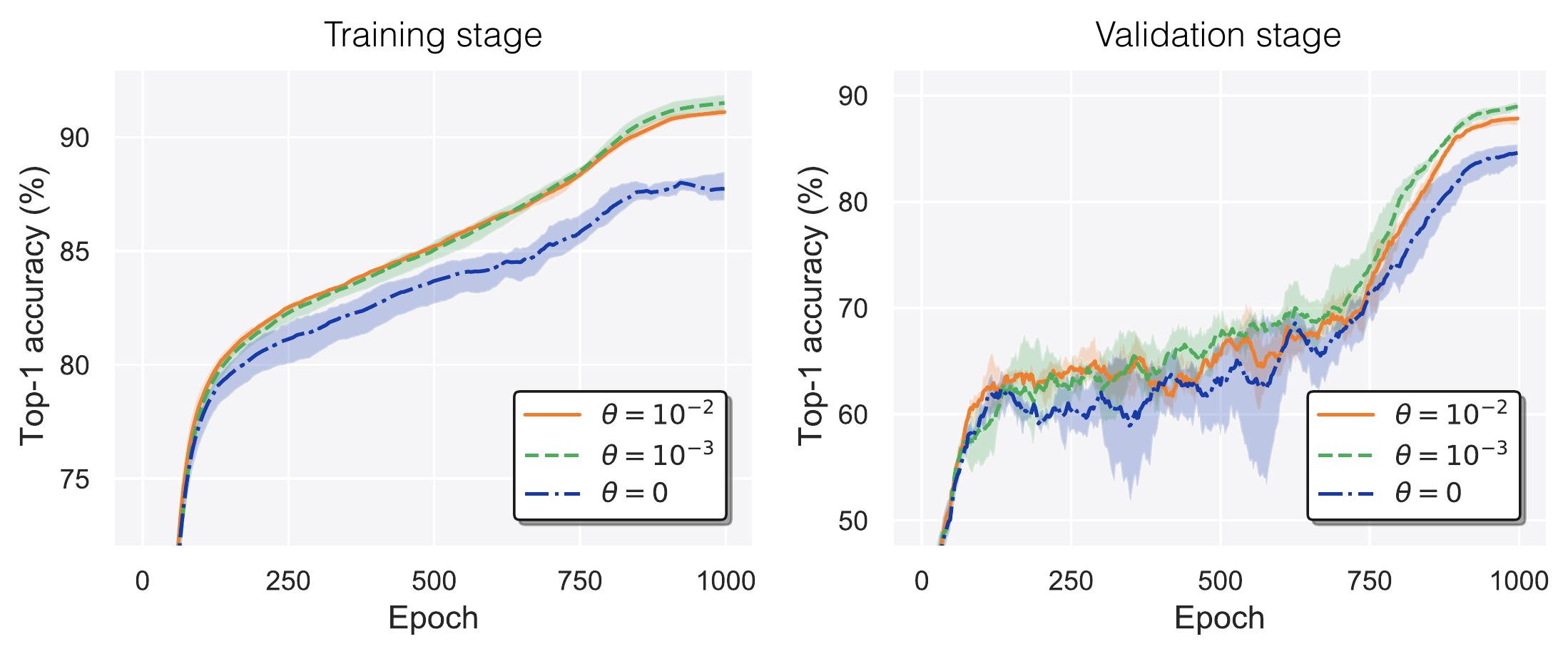}
\caption{Top-1 accuracies vs. epochs with different $\theta$ settings in $0.56$-bit SNNs, regarding the training stage (Left) and the validation stage (Right). Experiments are performed on CIFAR10 with ResNet-20, and each setting is performed three times with different random seeds.}
\label{pic:theta}
\vskip-0.06in
\end{figure}

\textbf{Analysis of hyper-parameter $\theta$.} In Eq.~(\ref{eq:mask}), we introduce $\theta$ as a threshold to alleviate the rapid sign change when $\rvp^i$ is around $0$ during training. To verify whether $\theta$ works, in Figure~\ref{pic:theta}, we provide the accuracy curves with different $\theta$ settings. When $\theta=0$, the validation curve oscillates during $100\hskip-0.02in\sim\hskip-0.02in 200$ epochs, and the  final prediction is impacted by the instability. We observe that setting $\theta$ to $10^{-3}$ or $10^{-2}$ leads to similar results, and $\theta=10^{-3}$ is slightly better.

\begin{table}[t]
	\centering
	\caption{Comparison of using layer-specific subsets and layer-shared subsets. Experiments are performed on  CIFAR10  with ResNet-20 and each setting is performed three times with different random seeds. Numbers highlighted in green are gains of the layer-specific settings over their layer-shared counterparts.
	} \vskip 0.05in
	\tablestyle{1pt}{1.4}
	\resizebox{1\linewidth}{!}{
		\begin{tabular}{c|c|ccccc}
			\label{tab:specific}
			\fontsize{8pt}{1em}\selectfont \multirow{2}*{Method}
			& \fontsize{8pt}{1em}\selectfont \multirow{2}*{Subsets strategy}
			
			& \multicolumn{4}{c}{\fontsize{8pt}{1em}\selectfont Bit-width (W$/$A) }
			\\
			&& \fontsize{8pt}{1em}\selectfont 0.11$/$32
			& \fontsize{8pt}{1em}\selectfont 0.33$/$32
			& \fontsize{8pt}{1em}\selectfont 0.56$/$32
			& \fontsize{8pt}{1em}\selectfont 0.78$/$32\\
			
			\shline
			
			Vanilla-SNN&Layer-shared & 70.2$\pm$6.5& 79.4$\pm$3.7& 82.6$\pm$2.2& 88.2$\pm$1.0 \\ 
			SNN&Layer-shared & 73.5$\pm$3.3& 82.5$\pm$2.6& 86.7$\pm$1.3& 90.0$\pm$0.3\\ 
			\cdashline{1-6}[1pt/1pt]
			
			Vanilla-SNN&Layer-specific & 80.1$\pm$2.3\cgaphl{+}{9.9} & 85.5$\pm$2.1\cgaphl{+}{6.1}& 87.8$\pm$1.6\cgaphl{+}{5.2} &  89.9$\pm$0.6\cgaphl{+}{1.7} \\
			SNN &Layer-specific & 81.6$\pm$1.7\cgaphl{+}{8.1}& 86.8$\pm$1.4\cgaphl{+}{4.3}& 88.9$\pm$0.4\cgaphl{+}{2.5}& 90.2$\pm$0.3\cgaphl{+}{0.2}\\
		\end{tabular}
	}
\end{table}

Our supplementary materials include training details, extension to $1\times1$ convolutions, evaluation on object detection with PASCAL VOC~\cite{ijcvEveringhamGWWZ10} and MS-COCO~\cite{eccvLinMBHPRDZ14} datasets, etc.

\subsection{Deployment Results}
\label{subsec:deployment_results}

\begin{table}[t]
	\centering
	\caption{Practical speed tests of  $0.56$-bit SNNs and $1$-bit BNNs deployments. The running time is evaluated with the $224\times 224$ image input and based on a hardware configuration of 64PEs@1GHz. Activations are not binarized in the models.} \vskip 0.1in
	\tablestyle{7pt}{1.1}
	\resizebox{0.8\linewidth}{!}{
		\begin{tabular}{c|cc|c}
			\label{tab:simulation}
			\fontsize{8pt}{1em}\selectfont \multirow{2}*{Backbone}
			& \multicolumn{2}{c|}{\fontsize{8pt}{1em}\selectfont Running time (ms)}
			& \fontsize{8pt}{1em}\selectfont \multirow{2}*{Speed up}
			\\
			&\fontsize{8pt}{1em}\selectfont $1$-bit BNN 
			& \fontsize{8pt}{1em}\selectfont $0.56$-bit SNN\\
			\shline
			ResNet-18&3.626 & \textbf{1.159}&{\fontsize{8pt}{1em}\selectfont \hl{\textbf{3.13}$\times$}} \\ 
			ResNet-34&7.753 & \textbf{2.329} &{\fontsize{8pt}{1em}\selectfont \hl{\textbf{3.33}$\times$}}
		\end{tabular}
	}
\vskip-0.1in
\end{table}

\begin{figure}[t!]
\centering
\hskip -0.2em
\includegraphics[scale=0.42]{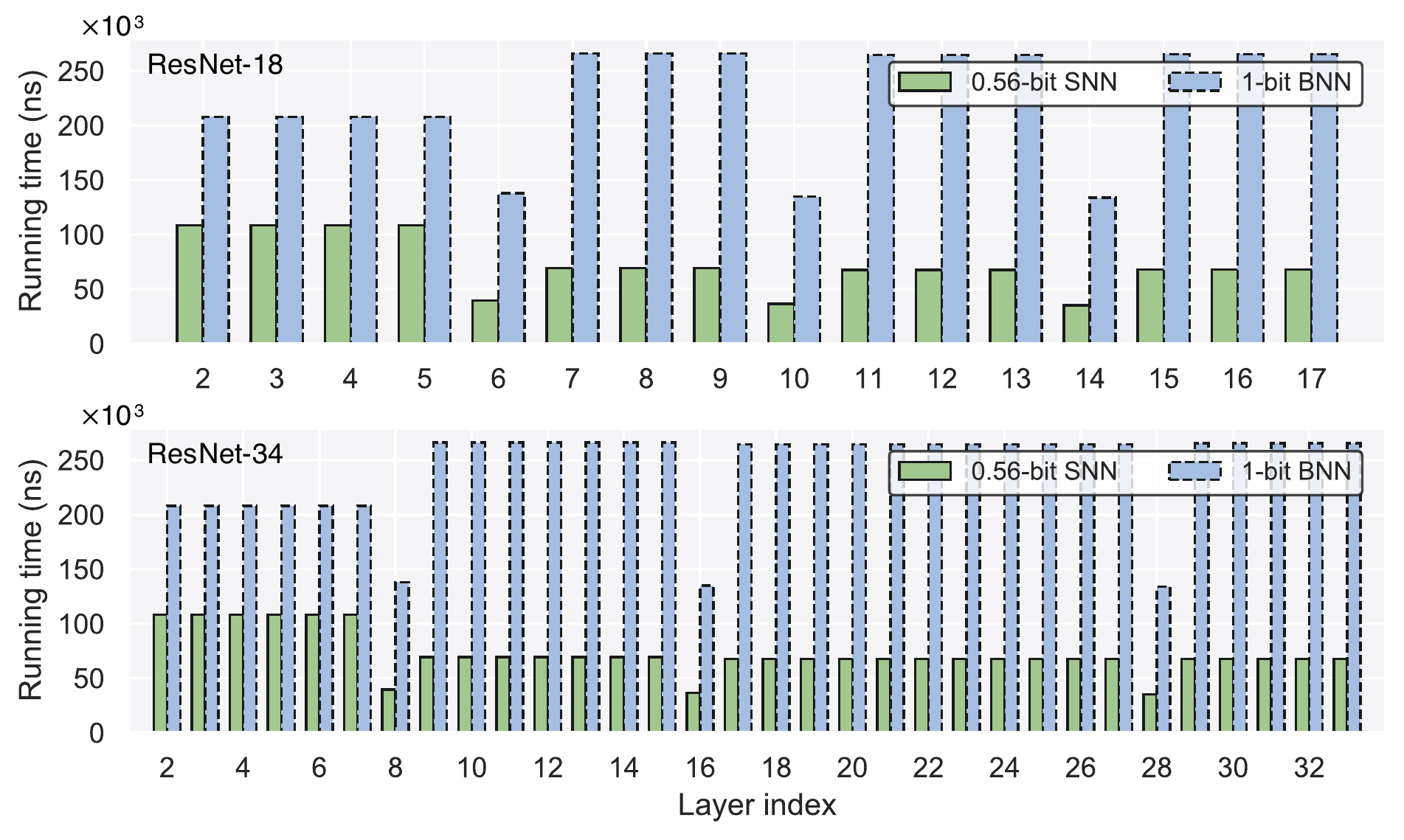}
\caption{Comparison of the running time per layer for $0.56$-bit SNNs and $1$-bit BNNs deployments.  }
\label{pic:layer_speed}
\vskip-0.05in
\end{figure}

We use Intel$^\circledR$ CoFluent$^\text{TM}$ Technology\footnote{\scriptsize{\url{https://www.intel.com/content/www/us/en/cofluent/overview.html}}}  to build a transaction level model and evaluate the performance of standard BNNs and our hardware design for SNNs.  Intel CoFluent is a system modeling and simulation solution for performance predicting and architecture optimization during early design stages. Results of ResNet-18 and ResNet-34 are provided in Table~\ref{tab:simulation}, and per layer comparison is illustrated in Figure~\ref{pic:layer_speed}. We observe that our SNNs constantly achieve faster speeds on all layers compared with their BNN counterparts. For either ResNet-18 or ResNet-34, our $0.56$-bit SNNs are \textbf{over three times faster} than the corresponding BNNs. In summary, compared with BNNs, when given the same computational resources, SNNs achieve much faster runtime speeds; when given the same throughput, SNNs greatly reduce the number of Arithmetic and Logic Units (ALUs) and thus reduce the chip area and power consumption.

\section{Conclusion} We propose Sub-bit Neural Networks (SNNs) which further compress and accelerate BNNs. The motivation  is inspired by the observation that binary kernels in BNNs are largely clustered, especially in large networks or deep layers. Our method consists of two steps, randomly sampling layer-specific subsets of binary kernels, and subsets refinement with optimization. Experimental results and practical deployments verify that SNNs can be remarkably efficient compared with BNNs while maintaining high accuracies.

\section*{Acknowledgement}
\vspace{0.03 in}
This work is funded by Major Project of the New Generation of Artificial Intelligence (No. 2018AAA0102900) and the Sino-German  Collaborative Research Project Crossmodal Learning (NSFC  62061136001/DFG TRR169). We thank Feng Chen and Zhaole Sun for the insightful discussions.

\clearpage 
\section*{\LARGE Appendix}
\appendix

\section{Training  Details}
For image classification experiments on both CIFAR10 and ImageNet datasets, networks are trained from scratch. We set the batch size to 256, and use an SGD optimizer with a momentum of 0.9.  The weight decay rate is assigned to  $10^{-4}$. We initialize the learning rate to 0.1 and adopt a cosine learning rate scheduler. Following IR-Net~\cite{cvprQinGLSWYS20}, we balance and standardize weights in the forward propagation and adopt the error decay estimator to approximate the $\mathrm{sign}$ function in the backward propagation. We select $\mathrm{Hardtanh}$ as the activation function instead of $\mathrm{ReLU}$ when we binarize activations~\cite{cvprQinGLSWYS20}. To networks considered in the experiments, we use standard data augmentation strategies following the original papers~\cite{ijcvLiuLWYLC20,eccvRastegariORF16, eccvLQNet}. 
On CIFAR10, we train each model on a single V100 GPU with 1000 epochs. On ImageNet, we train each model on four V100 GPUs with 120 epochs for ResNet-18 and ResNet-34. 

For object detection, on both PASCAL VOC and MS-COCO datasets, networks are pre-trained on ImageNet~\cite{Deng2009ImageNet} classification dataset. Following BiDet~\cite{cvprWangWL020}, the batch size is set to 32, and we adopt the Adam optimizer~\cite{DBLP:journals/corr/KingmaB14}. The learning rate is initialized to  0.001 which decays twice by multiplying 0.1 at the 6-th and 10-th epoch out of 12 epochs ~\cite{DBLP:journals/corr/KingmaB14}. We adopt four V100 GPUs for training.

\begin{figure}[t!]
\centering
\hskip -0.2em
\includegraphics[scale=0.398]{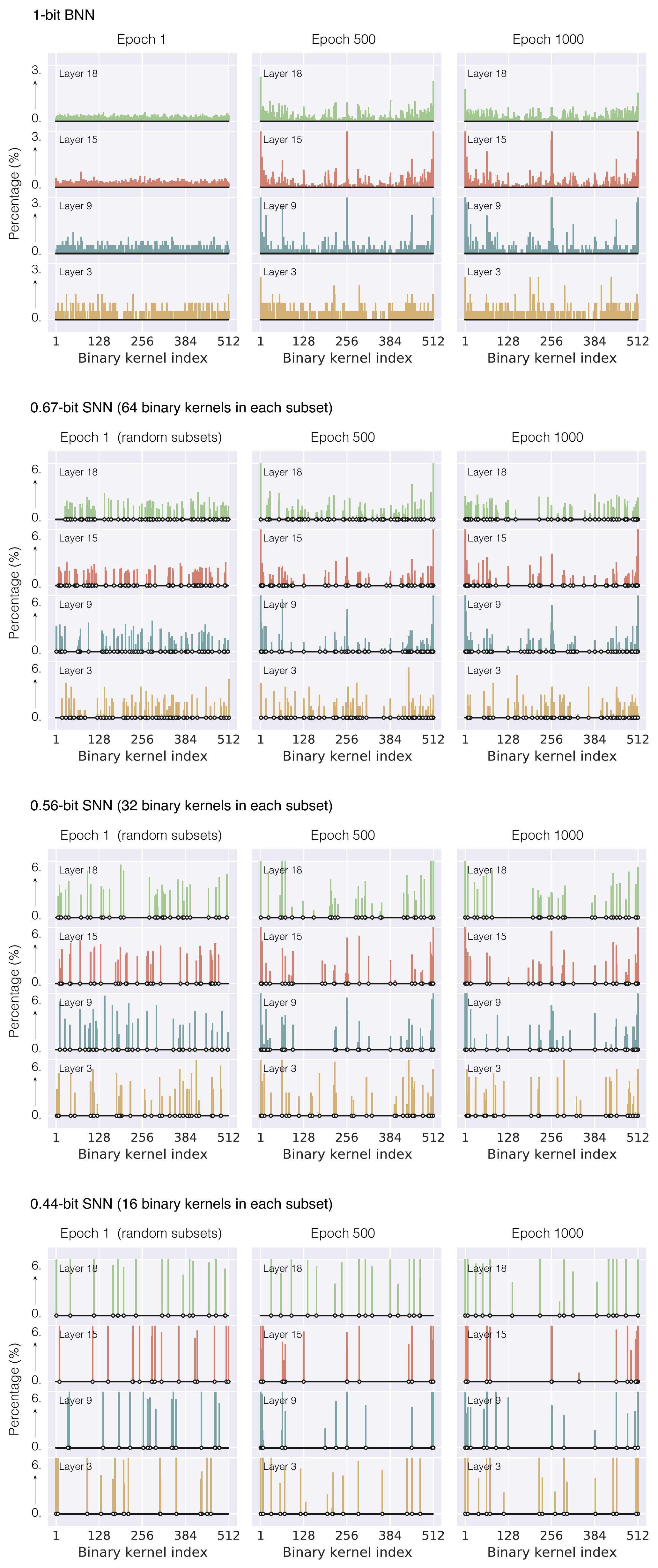}
\vskip0.05em
\caption{Visualization of how indices and frequencies of binary kernels change during training with subsets refinement. We provide results for $0.67$-bit, $0.56$-bit, and $0.44$-bit SNNs, as well as a $1$-bit BNN for comparison. Experiments are performed on CIFAR10 with ResNet-20. }
\label{pic:epoch_morebits}
\end{figure}

\section{More Visualizations}

To figure out how the subsets refinement module affects the binary kernels during training, we provide more visualization results in Figure~\ref{pic:epoch_morebits}. We observe that for different bit-width settings, binary kernels tend to be grouped and symmetric, especially in $0.56$-bit and $0.44$-bit subfigures. Besides, after training, binary kernels in SNNs present similar distributions in the same layers to some extent.

\section{Extending to Bottleneck with $1\hskip-0.03in \times\hskip-0.03in 1$ Kernels} 
In our main paper, we focus on $3\hskip-0.03in \times\hskip-0.03in 3$ kernels to design our SNNs, as $3\times$3 convolutional layers usually occupy major parameters and computational costs in many backbones such as VGG-small, ResNet-18, ResNet-34, etc. 
Yet in larger backbones like ResNet-50, the Bottleneck structure also contains $1\hskip-0.03in \times\hskip-0.03in  1$ convolutional layers  of which both the parameters and computations cannot be ignored. Therefore in this part, we introduce how our method can be naturally extended to such architectures.

For a $1\hskip-0.01in \times\hskip-0.01in 1$ convolutional layer, assuming its layer index is $i$, the weights can be denoted as $\rvw^i\in\mathbb{R}^{c^i_{out}\cdot c^i_{in}\times 1 \times 1}$. As in modern backbones  $c^i_{in}$ is usually divisible by $8$, we split $\rvw^i$ into $c^i_{out}\cdot\frac{ c^i_{in}}{8} $ vectors with each vector denoted as $\rvw^i_c\in\mathbb{R}^{8\times 1\times 1}$, where $c=1,2,\cdots,c^i_{out}\cdot\frac{ c^i_{in}}{8} $. In this case, binarizing $\rvw^i_c$ is formulated as $\bar{\rvw}^i_c=\argmin_{\rvk\in\mathbb{K}^{'}}\|\rvk-\rvw^i_c\|_2^2$, where $\mathbb{K}'=\{\pm1\}^{8\times1\times1}$ and $|\mathbb{K}^{'}|=256$. Again, we sample layer-specific subsets $\mathbb{P}'^i\hskip-0.2em\subset\hskip-0.2em\mathbb{K}^{'}$.  Under the circumstance, if we use $\tau'$-bit to represent a vector $\rvw^i_c$, there is $|\mathbb{P}'^i|=2^{\tau'}$. By using indices $1,2,3,\cdots,2^{\tau'}$ to represent each binarized vector $\bar{\rvw}^i_c$, we can obtain a compression ratio $\frac{\tau'}{8}$.

To facilitate the understanding, Table~\ref{tab:compare1x1} illustrates the comparison of the binarization for  $3\hskip-0.01in \times\hskip-0.01in 3$ kernels (proposed in our main paper) and for  $1\hskip-0.01in \times\hskip-0.01in 1$ kernels.

In Figure 6 of our main paper, we introduce a method that accelerates the $3\times3$ convolution operations of SNNs. $1\times1$ convolutional layers of SNNs can also be accelerated by the similar computation sharing.

Table~\ref{tab:resnet50} provides experimental results based on ResNet-50, and we again provide the full results for comparison. We adopt three bit-width settings including $0.59$-bit, $0.47$-bit, and $0.35$-bit, which correspond to setting both $\tau$ and $\tau'$ to $5,4,3$ respectively. We observe that the $0.47/32$-bit setting with ResNet-50 only drops $0.6\%$ in top-1 accuracy compared with the $1/32$-bit counterpart, yet it achieves $2.13\times$ parameter reduction and $5.27\times$ Bit-OPs reduction.

\begin{table}[t]
	\centering\vskip 0.05in
	\caption{Comparison of binarization methods including BNN and SNN for $3\times3$ and $1\times1$ convolutional layers, including weights representation, element of a set/subset, set/subset representation, bit numbers of $\rvw^i_c$ and per weight.
	} \vskip 0.05in
	\tablestyle{4pt}{1.3}
	\resizebox{1\linewidth}{!}{
		\begin{tabular}{ll|cc}
			\label{tab:compare1x1}
			&&\fontsize{8pt}{1em}\selectfont $3\times3$ kernel size
			& \fontsize{8pt}{1em}\selectfont $1\times1$ kernel size	\\
			\shline
			\multicolumn{2}{l|}{\fontsize{8pt}{1em}\selectfont {Weights}}&$\rvw^i\in\mathbb{R}^{c^i_{out}\cdot c^i_{in}\times 3 \times 3}$ & $\rvw^i\in\mathbb{R}^{c^i_{out}\cdot c^i_{in}\times 1 \times 1}$\\ 
			\multicolumn{2}{l|}{\fontsize{8pt}{1em}\selectfont {Element of a set/subset}} & $\rvw^i_c\in\mathbb{R}^{1\times 3\times 3} $ (Kernel)  &$\rvw^i_c\in\mathbb{R}^{8\times 1\times 1}$ (Vector) \\ 
			\multicolumn{2}{l|}{\fontsize{8pt}{1em}\selectfont {Number of units}} & $c^i_{out}\cdot c^i_{in}$ &$c^i_{out}\cdot\frac{ c^i_{in}}{8} $ \\ 
			\cdashline{1-4}[1pt/1pt]

			\fontsize{8pt}{1em}\selectfont \multirow{3}*{BNN} &Full set  & $\mathbb{K}=\{\pm1\}^{1\times 3\times 3}$ & $\mathbb{K}'=\{\pm1\}^{8\times 1\times 1}$\\
			&Bits of $\rvw^i_c$& $9$-bit & $8$-bit\\
			&Bits per weight& $1$-bit & $1$-bit\\
			\cdashline{1-4}[1pt/1pt]
			\fontsize{8pt}{1em}\selectfont \multirow{3}*{SNN} 
			& Subset  & $\mathbb{P}^i\hskip-0.2em\subset\hskip-0.2em\mathbb{K}$, $|\mathbb{P}^i|=2^\tau$ & $\mathbb{P}'^i\hskip-0.2em\subset\hskip-0.2em\mathbb{K}^{'}$, $|\mathbb{P}'^i|=2^{\tau'}$\\
			&Bits of $\rvw^i_c$& $\tau$-bit, $1\le\tau<9$ & $\tau'$-bit, $1\le\tau<8$\\
			&Bits per weight& $\frac{\tau}{9}$-bit & $\frac{\tau'}{8}$-bit\\
		\end{tabular}
	}
\vskip-0.02in
\end{table}

\begin{table}[t]
	\centering
	\caption{Results on the CIFAR10 dataset to verify our method on ResNet-50 which contains $1\times 1$ convolutional layers. Single-crop testing with $32 \times 32$ crop size is adopted, and each result of our method is the average of three runs. Numbers highlighted in green are reduction ratios over BNN counterparts. $^*$ indicates our implemented results.
	} \vskip 0.05in
	\tablestyle{4pt}{1}
	\resizebox{1\linewidth}{!}{
		\begin{tabular}{c|c|c|c|c}
			\label{tab:resnet50}
			\fontsize{8pt}{1em}\selectfont \multirow{2}*{Method} 
			& \fontsize{8pt}{1em}\selectfont Bit-width
			& \fontsize{8pt}{1em}\selectfont \#Params
			& \fontsize{8pt}{1em}\selectfont Bit-OPs
			& \fontsize{8pt}{1em}\selectfont Top-1 Acc.\\
			&\fontsize{8pt}{1em}\selectfont (W$/$A)
			&\fontsize{8pt}{1em}\selectfont  (Mbit)
			& \fontsize{8pt}{1em}\selectfont (G)
			& \fontsize{8pt}{1em}\selectfont (\%)
			\\
			\shline
			\multicolumn{5}{c}{\fontsize{8pt}{1em}\selectfont \modify{\textbf{ResNet-50} (Extending to Bottleneck with $1\hskip-0.03in \times\hskip-0.03in 1$ convolutional kernels)}}\\
			\cdashline{1-5}[1pt/1pt]
			Full precision  &32$/$32& 750.26& 78.12 & 95.4 \\
			\cdashline{1-5}[1pt/1pt]
			IR-Net$^*$~\cite{cvprQinGLSWYS20} & 1$/$1 &23.45 & 1.221& 93.2 \\
			\rowcolor{cyan!7}
			Vanilla-SNN $\vert$ SNN & 0.59$/$1 & 13.87\cgaphlsupp{1.7}{$\times$}&0.333\cgaphlsupp{3.7}{$\times$}& 92.1 $\vert$  92.9\\
			\rowcolor{cyan!7}
			Vanilla-SNN $\vert$ SNN & 0.47$/$1 & 11.09\cgaphlsupp{2.1}{$\times$}&0.239\cgaphlsupp{5.1}{$\times$}& 91.4 $\vert$ 92.4 \\
			\rowcolor{cyan!7}
			Vanilla-SNN $\vert$ SNN & 0.35$/$1 & 8.321\cgaphlsupp{2.8}{$\times$}&0.191\cgaphlsupp{6.4}{$\times$}& 91.0 $\vert$  92.1 \\
			\cdashline{1-5}[1pt/1pt]
			IR-Net$^*$~\cite{cvprQinGLSWYS20} & 1$/$32 & 23.45 & 39.06& 95.1 \\
			\rowcolor{cyan!7}
			Vanilla-SNN $\vert$ SNN & 0.59$/$32 & 13.87\cgaphlsupp{1.7}{$\times$}&10.67\cgaphlsupp{3.7}{$\times$}& 94.4 $\vert$ 95.1 \\
			\rowcolor{cyan!7}
			Vanilla-SNN $\vert$ SNN & 0.47$/$32 & 11.09\cgaphlsupp{2.1}{$\times$}&7.640\cgaphlsupp{5.1}{$\times$}& 93.8 $\vert$ 94.5 \\
			\rowcolor{cyan!7}
			Vanilla-SNN $\vert$ SNN & 0.35$/$32& 8.321\cgaphlsupp{2.8}{$\times$}&6.127\cgaphlsupp{6.4}{$\times$}& 93.5 $\vert$ 94.0  \\

		\end{tabular}
	}
\vskip -0.08in
\end{table}

\section{Evaluation on Object Detection}
To prove the generalization of our SNNs on object detection, we further perform experiments on PASCAL VOC~\cite{ijcvEveringhamGWWZ10} and MS-COCO 2014~\cite{eccvLinMBHPRDZ14} datasets. PASCAL VOC contains images within 20 different categories. The same with~\cite{cvprWangWL020}, we train our SNNs using VOC 2007 and VOC 2012 trainval-sets (16k images) and evaluate on VOC 2007 test-set (5k images). MS-COCO 2014 dataset contains images within 80 different categories. We follow the popular ``trainval35k'' and ``minival'' data split method~\cite{cvprBellZBG16,cvprWangWL020}, which sets up training with the combination of  the training set (80k images) as well as sampled images from the validation set (35k images), and  adopts the remaining 5k images in the validation set for testing. 

We choose two typical pipelines including SSD300~\cite{DBLP:conf/eccv/LiuAESRFB16} and Faster R-CNN~\cite{DBLP:conf/nips/RenHGS15}, using VGG16 and ResNet-18 as backbones respectively. Following the standard evaluation metrics~\cite{eccvLinMBHPRDZ14}, we report the average precision (AP) for IoU$\in[0.5\hskip-0.2em:\hskip-0.2em 0.05\hskip-0.2em :\hskip-0.2em 0.95]$, denoted as mAP, and $\text{AP}_{50}$, $\text{AP}_{75}$ as well. 

For a fair comparison, we follow the same training settings with BiDet~\cite{cvprWangWL020}, which achieves state-of-the-art binarization performance for the object detection task. Evaluation results on both datasets are reported in Table~\ref{tab:object_detection}, and we also consider $0.67$-bit, $0.56$-bit, and $0.44$-bit settings for our SNNs. Regarding Faster-RCNN on the PASCAL VOC dataset, reducing the bit-width to $0.56$, mAP has a slight drop of  $1.6\%$; using $0.67$-bit on the COCO dataset achieves very close performance compared with $1$-bit BiDet, with only $0.6\%$ mAP drop. These evaluation results again prove the effectiveness of our proposed method.

\begin{table}[t]
	\centering
	\caption{Object detection results on the PASCAL VOC and MS-COCO 2014 datasets. Single-crop testing with crop size $300 \times 300$ for SSD300, and $600 \times 1000$ for Faster R-CNN. We follow the training details and techniques in BiDet~\cite{cvprWangWL020}, which could be the reference for our $1$-bit baselines. 
	} \vskip 0.05in
	\tablestyle{3pt}{1.1}
	\resizebox{1\linewidth}{!}{
		\begin{tabular}{c|c|c|ccc}
			\label{tab:object_detection}
			\fontsize{8pt}{1em}\selectfont \multirow{2}*{Method} 
			& \fontsize{8pt}{1em}\selectfont Bit-width 
			& \fontsize{8pt}{1em}\selectfont VOC
			& \multicolumn{3}{c}{\fontsize{8pt}{1em}\selectfont MS-COCO 2014}
			\\
			
			& \fontsize{8pt}{1em}\selectfont (W$/$A)
			& \fontsize{8pt}{1em}\selectfont mAP (\%)
			& \fontsize{8pt}{1em}\selectfont mAP (\%)
			& \fontsize{8pt}{1em}\selectfont $\text{AP}_{50}$ (\%)
			& \fontsize{8pt}{1em}\selectfont $\text{AP}_{75}$ (\%) \\
			\shline
			\multicolumn{6}{c}{\fontsize{8pt}{1em}\selectfont {VGG16, SSD}}\\
			\cdashline{1-6}[1pt/1pt]
			Full precision &32$/$32& 72.4 &  23.2 & 41.2 & 23.4 \\
			\cdashline{1-6}[1pt/1pt]
			BNN~\cite{nipsHubaraCSEB16} & 1$/$1 & 42.0 & 6.2 & 15.9 & 3.8 \\
			XNOR~\cite{eccvRastegariORF16} &1$/$1& 50.2 &8.1 & 19.5 & 5.6 \\
			Bi-Real~\cite{ijcvLiuLWYLC20} &1$/$1& 63.8 & 11.2 & 26.0 & 8.3 \\
			BiDet~\cite{cvprWangWL020} &1$/$1& 66.0 & 13.2 & 28.3 & 10.5 \\
			\rowcolor{cyan!7}
			Vanilla-SNN $\vert$ SNN & 0.67$/$1 & 64.0 $\vert$ 65.1 & 12.1 $\vert$ 12.8 & 27.4 $\vert$ 27.9 &$\,$ 9.4 $\vert$ 10.1 \\
			\rowcolor{cyan!7}
			Vanilla-SNN $\vert$ SNN & 0.56$/$1 & 63.3 $\vert$ 64.2 &11.2 $\vert$ 11.9  & 26.3 $\vert$ 27.1 & 8.9 $\vert$ 9.5 \\
			\rowcolor{cyan!7}
			Vanilla-SNN $\vert$ SNN & 0.44$/$1 & 61.8 $\vert$ 62.9 & 10.1 $\vert$ 11.0& 24.9 $\vert$ 25.6 & 8.0 $\vert$ 8.7 \\
			\hline
			\multicolumn{6}{c}{\fontsize{8pt}{1em}\selectfont {ResNet-18, Faster R-CNN}}\\
			\cdashline{1-6}[1pt/1pt]
			Full precision &32$/$32&74.5 & 26.0 & 44.8  & 27.2 \\
			\cdashline{1-6}[1pt/1pt]
			BNN~\cite{nipsHubaraCSEB16} & 1$/$1 & 35.6 &5.6 & 14.3 & 2.6 \\
			XNOR~\cite{eccvRastegariORF16} &1$/$1& 48.4 & 10.4 & 21.6 & 8.8 \\
			Bi-Real~\cite{ijcvLiuLWYLC20} &1$/$1& 58.2 & 14.4 & 29.0 & 13.4 \\
			BiDet~\cite{cvprWangWL020} &1$/$1& 59.5 & 15.7 & 31.0 & 14.4 \\
			\rowcolor{cyan!7}
			Vanilla-SNN $\vert$ SNN & 0.67$/$1 & 57.6 $\vert$ 58.8 &14.3 $\vert$ 15.1 & 29.6 $\vert$ 30.5 & 13.3 $\vert$ 13.8\\
			\rowcolor{cyan!7}
			Vanilla-SNN $\vert$ SNN & 0.56$/$1 & 56.9 $\vert$ 57.9 & 13.5 $\vert$ 14.3  & 29.0 $\vert$ 29.9 & 12.4 $\vert$ 13.2 \\
			\rowcolor{cyan!7}
			Vanilla-SNN $\vert$ SNN & 0.44$/$1 & 55.2 $\vert$ 56.4 & 12.4 $\vert$ 13.3 & 27.7 $\vert$ 28.7 & 11.8 $\vert$ 12.5 
		\end{tabular}
	}
\vskip -0.1in
\end{table}

\section{Other Possible Baselines}
\label{sec:possiblemethods}
Besides the methods discussed and compared in the experiments of the main paper, we also consider another two possible methods as baselines. {1) Baseline A:} Since each binary kernel is assigned an index from $1$ to $512$, a straightforward solution is to simply select binary kernels uniformly with the same index interval. {2) Baseline B:} At every iteration or every 10 iterations, we select the top $2^\tau$ most frequent binary kernels from the corresponding $1$-bit BNNs as the subset for each layer. Results of both baselines and our method are provided in Table~\ref{tab:possiblemethods}. Here, experiments are conducted on ImageNet with ResNet-18.
We find that Baseline A achieves much lower performance than our proposed method, even Vanilla-SNN. We speculate that uniformly selecting binary kernels  impacts the network representation ability as subsets in different layers are the same. In contrast, Baseline B seems to be a stronger baseline and it also surpasses Vanilla-SNN. However, the counting and sorting processes slow down the training. The training speed issue of Baseline B could be alleviated by updating subsets every 10 training iterations instead of 1, but accordingly the accuracy slightly drops. These results verify that our  SNN is superior in both performance and training efficiency.

\begin{table}[t]
	\centering
	\caption{Results on the ImageNet dataset with ResNet-18 to verify two possible baseline methods described in Sec.~\ref{sec:possiblemethods}. ``1 iteration'' and ``10 iterations'' indicate updating binary kernels every iteration and every 10 iterations, respectively.
	} 
	\tablestyle{7pt}{1.1}
	\resizebox{0.75\linewidth}{!}{
		\begin{tabular}{c|c|c}
			\label{tab:possiblemethods}
			\fontsize{8pt}{1em}\selectfont \multirow{2}*{Method} 
			& \fontsize{8pt}{1em}\selectfont Bit-width
			& \fontsize{8pt}{1em}\selectfont Top-1 Acc.\\
			&\fontsize{8pt}{1em}\selectfont (W$/$A)
			& \fontsize{8pt}{1em}\selectfont (\%)
			\\
			\shline
			Baseline A & 0.56$/$32 & 60.5  \\
			Baseline B & 0.56$/$32 &  62.7  \\
			Baseline B-10 & 0.56$/$32 &  62.5  \\
			Vanilla-SNN $\vert$ SNN & 0.56$/$32 & 62.8 $\vert$  63.4 \\
			
		\end{tabular}
	}
\vskip -0.16in
\end{table}

\begin{figure}[t!]
\centering
\hskip -0.2em
\includegraphics[scale=0.39]{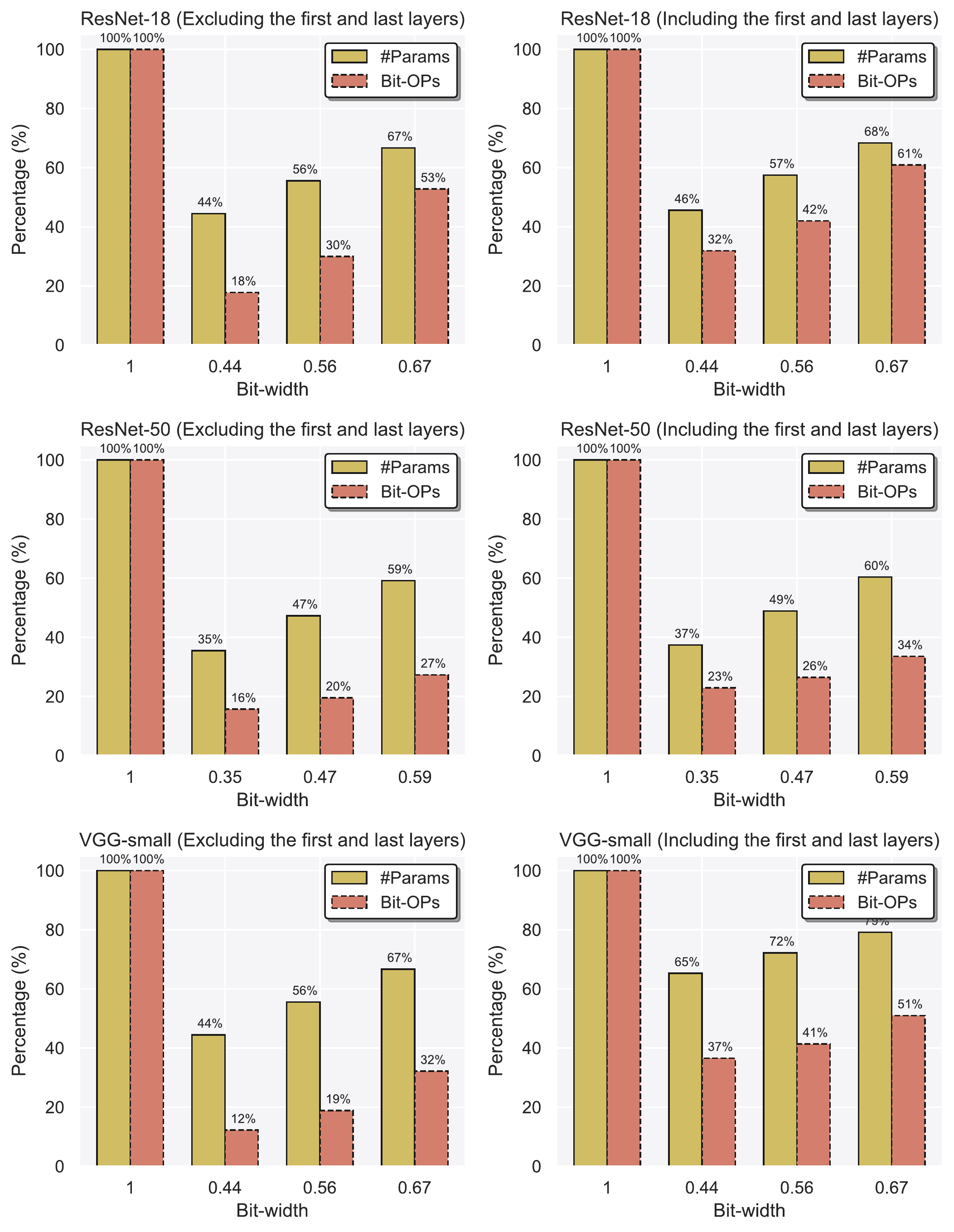}
\caption{Reduction ratios \emph{w.r.t.} parameters and bit-wise operations compared with $1$-bit BNNs. Subfigures in the left column provide the comparison when excluding the first and last layers. Subfigures in the right column provide the comparison when including the first and last layers with full-precision weights. Experiments are conducted on CIFAR10 with ResNet-18, ResNet-50, and VGG-small. Activations are not binarized in the models.}
\label{pic:including}
\vskip-0.02in
\end{figure}

\section{Compression and Acceleration Including the First and Last layers}

Regarding the common paradigm~\cite{ijcvLiuLWYLC20,cvprQinGLSWYS20,eccvRastegariORF16} of network binarization, weights in the first layer and last layer are full-precision weights, and in other layers are binarized. Our comparisons for parameters and bit-wise operations so far have excluded the first and last layers. To further verify the efficiency of our method when considering the whole network, Figure~\ref{pic:including} provides the comparison when excluding and including the first and last layers. We observe that our method can as well achieve good compression and acceleration performance even taking the first and last layers (with full-precision weights) into consideration. Note that VGG-small does not have a global average pooling layer, and thus its fully connected layer occupies more parameters and operations than the other two backbones.

\begin{figure}[t!]
\centering
\hskip -0.15em
\includegraphics[scale=0.32]{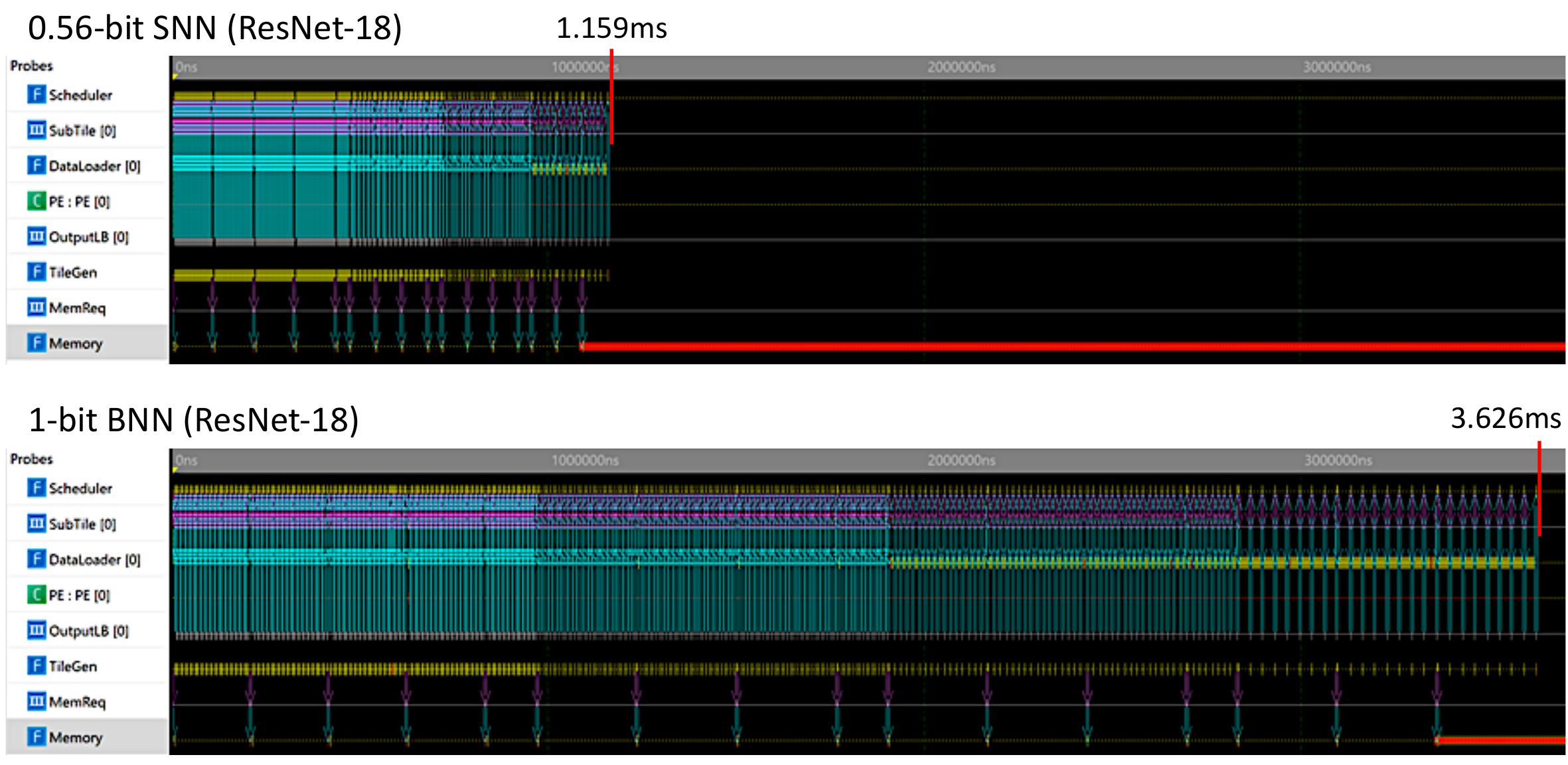}
\vskip0.2em
\caption{Timeline comparison between a $0.56$-bit SNN and a $1$-bit BNN, based on ResNet-18. This visualization is a supplement to Table 4 of our main paper. The running time is evaluated with the $224\times 224$ image input, and a hardware configuration of Intel$^\circledR$ CoFluent$^\text{TM}$ 64PEs@1GHz. Zoom in for the best view.}
\label{pic:timechart}
\vskip-0.06in
\end{figure}

\section{Other Details of Practical Deployment} 

In Sec.~\ref{subsec:deployment_results} of our main paper, we conduct practical deployment and  compare the total running time of $0.56$-bit SNNs and $1$-bit BNNs. Here, we provide Figure~\ref{pic:timechart} which depicts the evaluation timelines of both models based on ResNet-18. Different timeline densities  indicate different stages of ResNet-18. We observe that our SNN can achieve high acceleration ratios in the last three stages where the channel numbers are large, which is consistent with our analysis in Sec.~\ref{subsec:accelerate} of the main paper.

{\small
\bibliographystyle{ieee_fullname}
\balance
\bibliography{egbib}
}

\end{document}